\documentclass{article}

\usepackage{microtype}
\usepackage{graphicx}
\usepackage{booktabs} 

\usepackage{hyperref}

\usepackage{xspace}
\usepackage{amsmath}
\usepackage{multirow}
\usepackage[nounderscore]{syntax}
\usepackage[inline]{enumitem}
\usepackage{listings}
\usepackage{float}
\usepackage{multicol}
\usepackage[dvipsnames,svgnames]{xcolor}

\definecolor{codegreen}{rgb}{0,0.6,0}
\definecolor{codegray}{rgb}{0.5,0.5,0.5}
\definecolor{codepurple}{rgb}{0.58,0,0.82}
\definecolor{backcolour}{rgb}{0.95,0.95,0.92}

\lstdefinestyle{mystyle}{
    backgroundcolor=\color{backcolour},
    commentstyle=\color{codegreen},
    keywordstyle=\color{magenta},
    numberstyle=\tiny\color{codegray},
    stringstyle=\color{codepurple},
    basicstyle=\ttfamily\footnotesize,
    breakatwhitespace=false,
    breaklines=true,
    captionpos=b,
    keepspaces=true,
    numbers=left,
    numbersep=5pt,
    showspaces=false,
    showstringspaces=false,
    showtabs=false,
    tabsize=2
}
\usepackage{subcaption}
\usepackage{caption}

\usepackage[accepted]{mlsys2024}

\mlsystitlerunning{\xg: Flexible and Efficient Structured Generation Engine For Large Language Models}

\newif\ifdraft
\drafttrue 

\newcommand{\xg}{XGrammar\xspace}
\newcommand{\figref}[1]{Figure~\ref{#1}}
\newcommand{\secref}[1]{\S\ref{#1}}
\newcommand{\appref}[1]{Appendix~\ref{#1}}
\newcommand{\tableref}[1]{Table~\ref{#1}}
\newcommand{\algoref}[1]{Algorithm~\ref{#1}}

\newcommand{\myparagraph}[1]{\vspace{0pt}\paragraph{\textbf{#1}}}

\renewcommand{\algorithmiccomment}[1]{\{#1\}}

\begin{document}

\twocolumn[
\mlsystitle{\xg: Flexible and Efficient Structured Generation Engine For Large Language Models}



\mlsyssetsymbol{equal}{*}

\begin{mlsysauthorlist}
\mlsysauthor{Yixin Dong}{cmu}
\mlsysauthor{Charlie F. Ruan}{cmu}
\mlsysauthor{Yaxing Cai}{nvidia}
\mlsysauthor{Ruihang Lai}{cmu}
\mlsysauthor{Ziyi Xu}{sjtu}
\mlsysauthor{Yilong Zhao}{ucb}
\mlsysauthor{Tianqi Chen}{cmu,nvidia}
\end{mlsysauthorlist}

\mlsysaffiliation{cmu}{Carnegie Mellon University}
\mlsysaffiliation{ucb}{University of California, Berkeley}
\mlsysaffiliation{sjtu}{Shanghai Jiao Tong University}
\mlsysaffiliation{nvidia}{NVIDIA}

\mlsyscorrespondingauthor{Yixin Dong}{yixind@andrew.cmu.edu}

\mlsyskeywords{Machine Learning, MLSys}

\vskip 0.3in

\begin{abstract}
The applications of LLM Agents are becoming increasingly complex and diverse, leading to a high demand for structured outputs that can be parsed into code, structured function calls, and embodied agent commands. These developments bring significant demands for structured generation in LLM inference. Context-free grammar is a flexible approach to enable structured generation via constrained decoding. However, executing context-free grammar requires going through several stack states over all tokens in vocabulary during runtime, bringing non-negligible overhead for structured generation. In this paper, we propose \xg{}, a flexible and efficient structure generation engine for large language models. \xg{} accelerates context-free grammar execution by dividing the vocabulary into context-independent tokens that can be prechecked and context-dependent tokens that need to be interpreted during runtime. We further build transformations to expand the grammar context and reduce the number of context-independent tokens. Additionally, we build an efficient persistent stack to accelerate the context-dependent token checks. Finally, we co-design the grammar engine with LLM inference engine to overlap grammar computation with GPU executions. Evaluation results show that \xg{} can achieve up to 100x speedup over existing solutions. Combined with an LLM inference engine, it can generate near-zero overhead structure generation in end-to-end low-LLM serving.

\end{abstract}
]



\printAffiliationsAndNotice{}  

\section{Introduction}

\begin{figure*}[t]
    \centering
    \includegraphics[width=\linewidth]{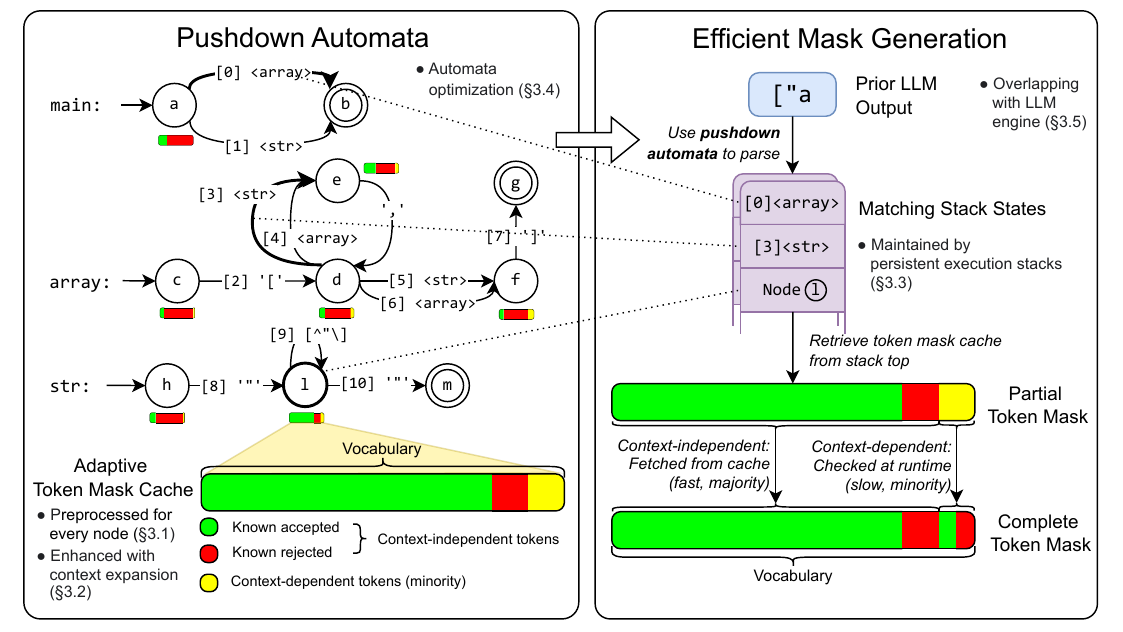}
    \caption{Overview of our approach. \xg first uses a pushdown automaton to parse the prior LLM output, flexibly supporting diverse grammars and producing the matching stack states. It then uses the stack top to index into the adaptive token mask cache—our key optimization—to retrieve a partial mask. Most of the partial mask consists of context-independent tokens and is determined during preprocessing. A small portion, however, is context-dependent and resolved at runtime. This yields the complete token mask, thus enabling efficient constraint decoding. }
    \label{fig:overview}
\end{figure*}

Recent advancements in large language models (LLMs) have created new possibilities for complex applications such as code generation~\cite{chen2021evaluatinglargelanguagemodels, wang2021codet5identifierawareunifiedpretrained}, debugging~\cite{pearce2022examiningzeroshotvulnerabilityrepair, mozannar2024reading}, external tool invocation through function calling~\cite{openai2024funccall, langchainToolCalling}, and robotic control~\cite{liu2023llmpempoweringlargelanguage}. These applications bring great demand for the \emph{structured generation} problem, which requires the output of LLMs conforms to specific formats or grammars, such as JSON, SQL or other formats tailored to the task. The downstream applications can then organically consume the structured outputs to perform followup interactions with the system.

Constrained decoding \cite{deutsch-etal-2019-general, kuchnik2023validating} is a commonly adopted method for structured generation. It guarantees the output of LLMs adheres to the specified structure through only allowing tokens that conform to the structure to be generated at each decoding step. At each step, constrained decoding first scans the entire vocabulary to identify invalid tokens and sets their probabilities to zero, thereby preventing them from being generated. To support the rich structure formats arising in diverse applications, a flexible mechanism is needed to specify and check the constraints. Context-free grammar (CFG)~\cite{1056813, poesia2022synchromesh, scholak-etal-2021-picard} provides a general approach for defining structures through a set of rules. Each rule contains a sequence of characters or other rules, allowing recursive composition to represent complex structures. Compared to alternative formats such as regular expressions, CFGs offer greater flexibility by allowing recursive structures, making them suitable for describing common languages such as JSON, SQL, and domain-specific languages (DSLs).

However, naively applying CFG to constrained decoding is not efficient because of its flexible nature. First, each decoding step needs to interpret CFG for every possible token in the vocabulary, which can be as large as 128k in Llama 3.1~\cite{dubey2024llama3herdmodels}. Additionally,  CFG interpretation requires a stack state that tracks the recursive rules matched so far, making it impossible to precompute and cache all combinatorial combinations of stack patterns ahead of time. Finally, each token in the LLM generation comprises multiple characters, which may cross the boundaries of grammar elements and cause further recursion or stack pop during runtime execution. The misaligned boundaries bring the need to handle them carefully during grammar execution.

In this paper, we introduce \xg{}, a flexible and efficient structured generation engine for large language models to address the above challenges. \xg{} builds a byte-level pushdown automaton to represent context-free grammars (CFGs). Our main insight(shown in \figref{fig:overview}) is to categorize the tokens into  \textbf{context-independent} tokens that can be decided only from the local context of automata and \textbf{context-dependent} tokens that require the entire stack state. We precompute the token correctness for all context-independent tokens and store them in an adaptive token mask cache with specific storage formats tailored to each automata location. We also build algorithms to expand the context of each local rule and reduce the number of context-dependent tokens. Additionally, we build a persistent stack-based system to enable rapid state branching and rollback, expediting context-dependent token checks and cache preprocessing. Finally, we co-designed the grammar engine with LLM inference engines to overlap the grammar computations with GPU computations, bringing minimal overhead for structured generation.

Evaluation shows that \xg{} can achieve up to 100x reduction in per-token latency for context-free grammar compared to current state-of-the-art methods. Additionally, the \xg{}-integrated LLM serving engine for Llama-3.1 models achieves up to an 80x speedup in end-to-end LLM serving with structured output on the H100 GPU. We are open-sourcing \xg{} and integrating it into major open-source LLM frameworks.

The main contribution of this paper is as follows:

\begin{itemize}
    \item We introduce an adaptive token mask cache that leverages context-independent tokens and significantly reduces mask generation overhead.
    \item We design a persistent execution stack that enables fast rollback operations, rapid state branching, and
rollback, expediting context-dependent token processing.
    \item We built an efficient grammar engine co-designed with the LLM serving framework to achieve minimal structured generation overhead.
\end{itemize}

\section{Background}
\label{sec:background}

\subsection{LLM Constrained Generation}

\begin{figure}[t]
    \centering
    \includegraphics[width=0.99\columnwidth]{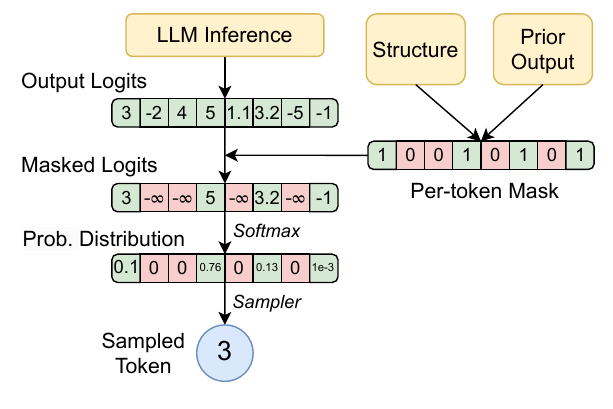}
    \caption{Constrained decoding with per-token mask. The per-token mask prevents LLM from generating tokens that would be invalid according to the structure at that step.}
    \label{fig:old_method}
\end{figure}

Large Language Models (LLMs) like GPT-4 \cite{openai2024gpt4technicalreport}, Llama \cite{dubey2024llama3herdmodels}, and Mistral \cite{jiang2023mistral7b} generate text in an auto-regressive manner, predicting one token at a time based on preceding sequence of tokens. The process starts with an initial prompt and continues as the model iteratively appends tokens until the response is complete. In LLMs, tokens serve as the basic input and output units. Each token represents a fixed string but may not correspond to a complete semantic unit or may break a Unicode character \cite{wang2019neuralmachinetranslationbytelevel}, creating challenges for structured text generation. At each step, the model produces a logits vector across its vocabulary, which is then converted into a probability distribution using the softmax function~\cite{NIPS1989_0336dcba}. A sampler then selects the next token from this distribution.

Constrained decoding guides the structure of LLM-generated text by restricting available tokens at each step, as illustrated in \figref{fig:old_method}. At each step, tokens that would violate the required structure are identified as invalid. Their logits are set to $-\infty$, effectively assigning them zero probability after the softmax operation and preserving the relative probabilities of other valid tokens. This ensures that only valid tokens are sampled. Efficiently identifying and masking invalid tokens is essential, as it directly impacts generation speed.

\begin{figure}[t]
    \centering
    \includegraphics[width=0.88\columnwidth]{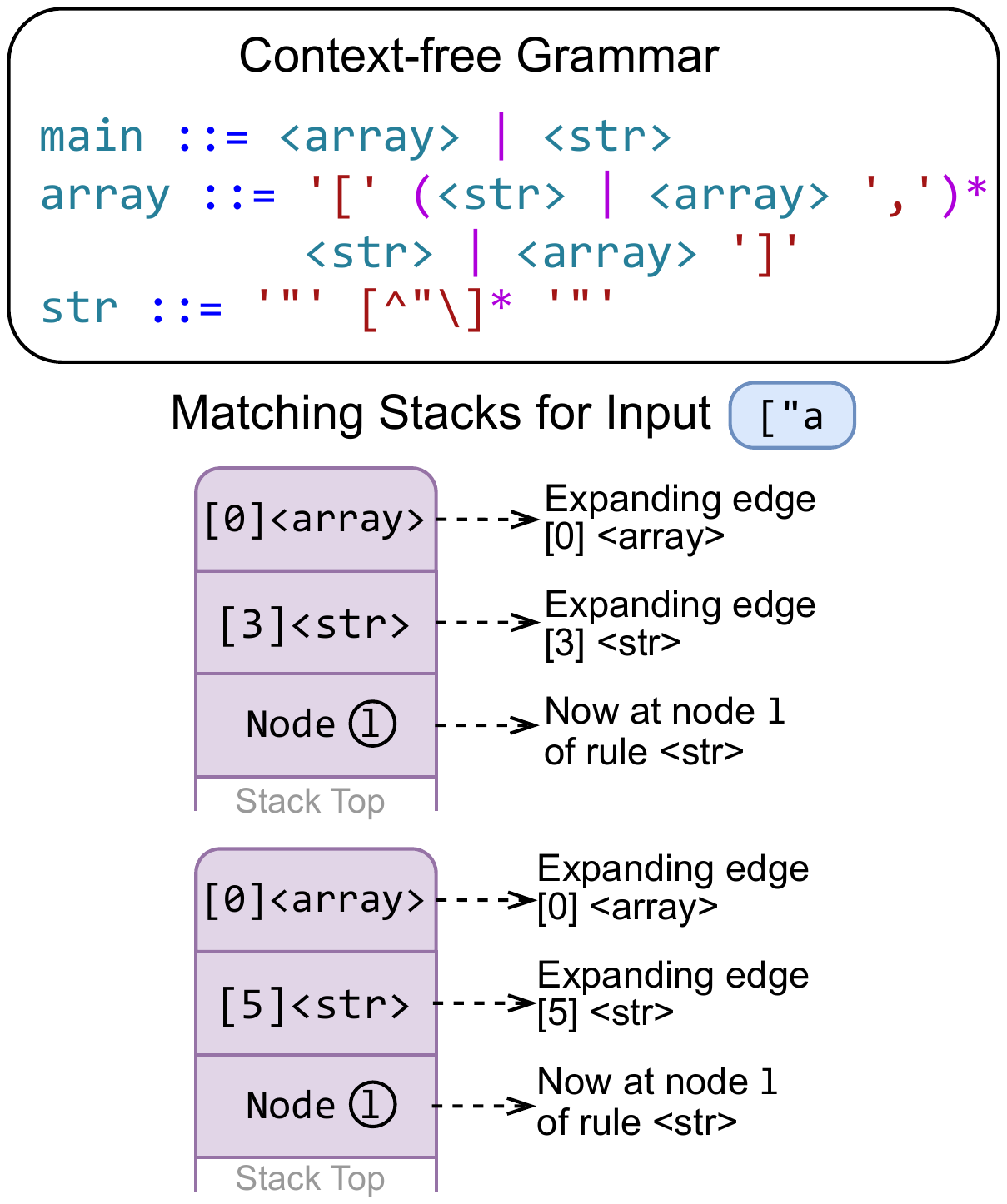}
    \caption{Up: A context-free grammar for arrays and strings that can be recursively composed. This CFG is converted into the pushdown automata in \figref{fig:overview}. \lstinline{[^"\\]} denotes every character except \lstinline{"} and \lstinline{\\}. Down: Two possible matching stacks for matching the string \lstinline{["a} to the CFG. Each stack represents a possible expansion of the rules in the CFG. The edges and nodes in the stack correspond to the transitions and states in the PDA in \figref{fig:overview}.}
    \label{fig:cfg}
\end{figure}




\subsection{Context-free Grammar and Pushdown Automata}

Context-free grammar (CFG) \cite{1056813} is widely used to define structures in structured generation. With an example shown in \figref{fig:cfg}, CFG contains multiple rules, each including characters or references to other rules, allowing recursive composition to define complex structures. This makes CFG suitable for languages such as JSON, SQL, and various domain-specific languages. CFG's recursive nature provides greater expressive power than simpler patterns, such as regular expressions, which are also frequently applied in LLM structured generation.

Pushdown automata (PDA)~\cite{SCHUTZENBERGER1963246, evey1963theory} are typically used to recognize languages generated by CFGs, as they employ a stack to manage nested structures. In this paper, we use a definition of PDA that is equivalent to the original one, but more conducive to explaining the algorithm. An example of PDA is shown in \figref{fig:overview}, and its stacks are shown in detail in \figref{fig:cfg}. A PDA consists of multiple finite state automata (FSA), each representing a grammar rule, with the stack handling recursive rule expansions. The transitions in the FSA include two types: character edges, which accept specific characters, and rule reference edges, which allow recursive entry into other rules. A formal definition of the PDA is provided in \appref{sec:appendix-pda}. To match a string, the PDA begins with the main rule, recursively expanding child rules by pushing rule-reference edges onto the stack; once a rule is fully matched, it pops the stack to return to the previous rule. The top of the stack holds the current node reached. If the grammar is non-deterministic, meaning there can be multiple possible transitions in the PDA for the same input character,  the PDA can maintain multiple parallel stacks for each path, ensuring flexibility. However, the unbounded stack length results in an infinite number of possible states, making it impractical to precompute token masks for all scenarios, thus posing challenges for efficient constrained decoding.

\section{\xg{}}

As shown in \figref{fig:overview}, \xg{} utilizes a byte-level pushdown automaton to interpret the context-free grammar. This byte-level design allows each character edge to include one or more bytes, handling irregular token boundaries and supporting tokens containing sub-UTF8 characters. The automaton's structure is optimized to accelerate matching, as described in \secref{sec:automata_optim}. In the preprocessing phase, we generate an adaptive token mask cache, as detailed in \secref{sec:token_mask_cache}, which accelerates runtime mask generation by precomputing context-independent tokens. The effectiveness of this cache is further enhanced by context extension in \secref{sec:context_expansion}. At runtime, the token mask cache quickly generates most of the mask, while the persistent execution stack in \secref{sec:persistent_stack} efficiently processes the rest context-dependent tokens. Additionally, mask generation and LLM inference are overlapped in \secref{sec:overlapping} to minimize the overhead of constrained decoding. Once the LLM generates a new token under the mask constraint, this token is then used to update the stack state of the pushdown automaton for the next mask generation.

\subsection{Adaptive Token Mask Cache}
\label{sec:token_mask_cache}

\begin{figure}[t]
    \centering
    \includegraphics[width=\columnwidth]{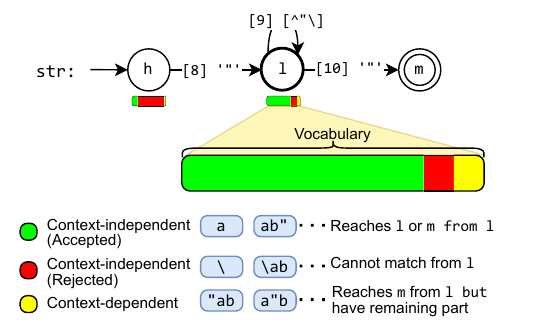}
    \caption{An example for the token mask cache. Tokens are categorized into three types: context-independent (accepted), context-independent (rejected), and context-dependent. The first two types can be directly determined for mask generation at runtime.}
    \label{fig:cache}
\end{figure}
To accelerate the generation of the token mask cache, the adaptive token cache categorizes tokens into two types (\figref{fig:cache}): context-independent tokens, which constitute the vast majority and can be pre-computed, and context-dependent tokens, which require slower, on-the-fly processing but are relatively few. This token classification relates to how tokens are validated by the pushdown automaton. We found that, considering the transition of the stack state, the process of matching tokens to the automaton can be divided into three categories:

\begin{enumerate}
    \item The matching process expands into a child rule, pushing new elements onto the stack.
    \item The matching process advances within the current rule, updating the stack top node to a new position.
    \item The matching process reaches the end of the current rule and returns to a parent rule, popping elements from the stack.
\end{enumerate}

Validating tokens in the former two cases only relies on the stack top node, which represents the position within the current rule, so we define these tokens as \emph{context-independent tokens}. The tokens in the third type, however, requires inspecting the entire running stack in validation, and are defined as \emph{context-dependent tokens}. For every node of the pushdown automaton, there is a set of context-independent tokens with this node being at the top of the stack at runtime, and their validity can be determined ahead of time. Therefore, we precompute the validity of these tokens and store them in a cache with the stack top node as the key, which we refer to as the adaptive token mask cache. It also adaptively selects the most efficient storage format based on the cache's contents, as explained in the next paragraph.

At runtime, we retrieve the validity of context-independent tokens directly based on the top of the stack to generate the token mask. The remaining few context-dependent tokens are validated by executing the pushdown automaton with the full stack. If parallel stacks exist due to the ambiguity of the grammar, the token masks for every stack is merged into a final token mask by finding the union of the accepted tokens in each mask. The computation for the token mask is significantly reduced because our method do not need to check context-independent tokens at runtime. Experiments show that context-dependent tokens account for only a minor proportion, amounting to less than 1\% (1134 out of 128k) for the Llama-3.1 model using JSON grammar.

\myparagraph{Adaptive storage.}

\begin{figure}[t]
    \centering
    \includegraphics[width=0.9\columnwidth]{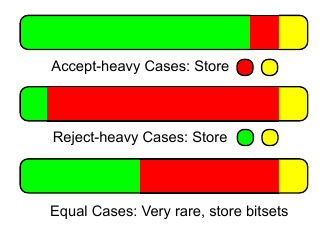}
    \caption{The adaptive storage format. In accept-heavy cases, we store the rejected tokens and context-dependent tokens. In reject-heavy cases, we store the accepted tokens and context-dependent tokens. In rare cases where two kinds of tokens are equal, we compress the accepted and rejected tokens into a bitset of the vocabulary size.}
    \label{fig:store}
\end{figure}

The token mask cache adopts an adaptive storage format to reduce memory usage, as illustrated in \figref{fig:store}. For each automaton node, the token mask cache divides the vocabulary into three parts: the accepted context-independent tokens, the rejected context-independent tokens, and the context-dependent tokens. Since these three parts together cover all tokens, it is sufficient to store only the two smaller subsets. We observe that, for a set of context-independent tokens, they tend to be either almost entirely accepted, namely \emph{accept-heavy} cases, or almost entirely rejected, namely \emph{reject-heavy} cases. This arises because, if wildcards can be matched from the current node, such as the wildcard \lstinline{[^"\]*} in the rule of string, nearly all tokens are valid; whereas if the node only accepts a few specific characters, nearly all tokens are invalid. Based on this observation, we designed the following adaptive storage format:
\begin{enumerate}
    \item For accept-heavy cases, we store the rejected context-independent tokens and context-dependent tokens in two arrays.
    \item For reject-heavy cases, we store the accepted context-independent tokens and context-dependent tokens in two arrays.
    \item For rare cases where the accepted and rejected tokens are roughly equal, we store the accepted and rejected context-independent tokens and compress them into a bitset matching the vocabulary size.
\end{enumerate}

Thus, in both accept-heavy and reject-heavy cases, the adaptive storage format only requires storing a small subset of tokens, significantly reducing memory usage. In practice, we will enumerate the three storage types, calculate their respective costs, and choose the storage type with the smallest size. For Llama-3.1 model and JSON grammar, this adaptive storage method can effectively reduce the total memory usage to 0.2\% (from 160 MB to 0.46 MB).

Additionally, when multiple parallel stacks exists, we need to merge the token masks. The merging algorithm of token masks is optimized based on storage type, as shown in \algoref{alg:merge_mask}. For an accept-heavy mask (many accepted tokens, storing only rejected tokens), it intersects the rejected tokens with $PartialRej$. For a reject-heavy mask (many rejected tokens, storing only accepted tokens), it combines accepted tokens with $PartialAcc$. In the final mask, the rejected tokens are the set difference $PartialRej \setminus PartialAcc$. This algorithm limits set operations to small token subsets, thus enhancing efficiency.

\begin{algorithm}
\caption{Efficiently Merge Token Masks}
\label{alg:merge_mask}
\begin{algorithmic}
\STATE {\bfseries Input:} Token masks for $k$ parallel stacks $\{M_i = (Acc_i, Rej_i)\}_{i=1}^k$, vocabulary $\mathcal{V}$.
\STATE {\bfseries Output:} The final token mask $M = (Acc, Rej)$.
\vspace{0.25em}
\STATE {\bfseries Initialize} $PartialAcc \gets \emptyset$, $PartialRej \gets \mathcal{V}$

\vspace{0.25em}
\FOR{$i=1$ {\bfseries to} $k$}
    \IF{$M_i$ is accept-heavy}
        \STATE $M_i$ only stores rejected token list $Rej_i$
        \STATE $PartialRej \gets PartialRej \cap Rej_i$
    \ELSE
        \STATE $M_i$ only stores accepted token list $Acc_i$
        \STATE $PartialAcc \gets PartialAcc \cup Acc_i$
    \ENDIF
\ENDFOR

\vspace{0.25em}

\STATE $M \gets \begin{aligned}[t]
&(\mathcal{V} \setminus (PartialRej \setminus PartialAcc), \\
&\quad PartialRej \setminus PartialAcc)
\end{aligned}$

\end{algorithmic}
\end{algorithm}

\subsection{Context Expansion}
\label{sec:context_expansion}

\begin{figure}[t]
    \centering
    \includegraphics[width=\columnwidth]{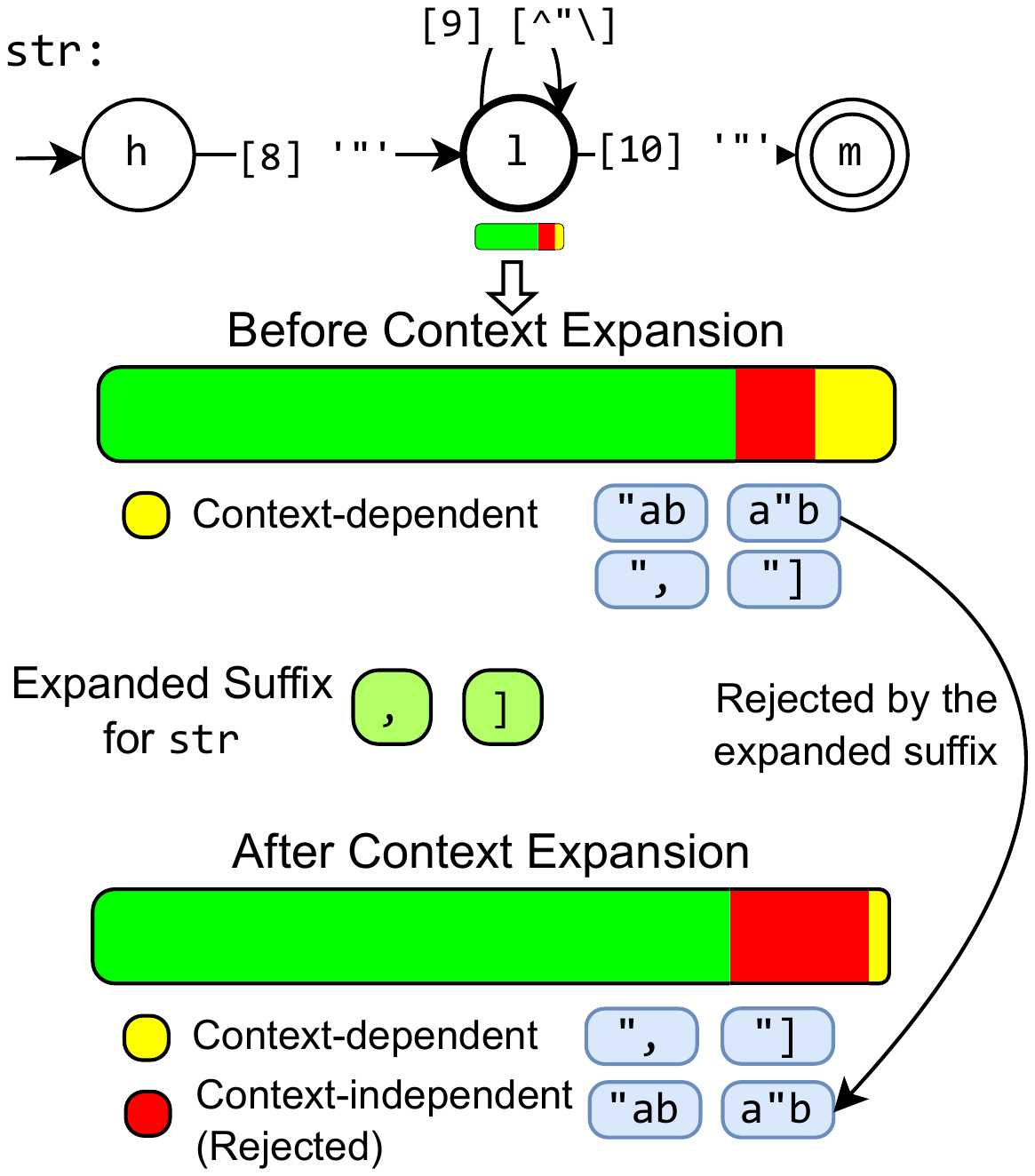}
    \caption{The context expansion. Each rule obtains a set of expanded suffices, representing the set of strings that must be matched after completing this rule. The remaining unmatched part of the context-dependent tokens should either be a prefix of the expanded suffix or start with the expanded suffix. Otherwise, they are rejected.}
    \label{fig:ctx_exp}
\end{figure}

\begin{algorithm}[tb]
\caption{Extract the Expanded Suffix Automaton}
\label{alg:context_expansion}
\begin{algorithmic}

\vspace{0.25em}
\STATE {\bfseries Input:} Pushdown automaton $\mathcal{P}$, rule $R$
\STATE {\bfseries Output:} Expanded context FSA $\mathcal{A}_R^{\text{ctx}}$ for $R$
\STATE {\bfseries Initialize} $\mathcal{A}_R^{\text{ctx}}$ as an empty FSA

\FOR{edge $s \xrightarrow{R} t$ in $\mathcal{P}$ referencing $R$}
    \STATE \COMMENT{ $\mathcal{A}_{\delta}$ is an FSA for the partial result }
    \STATE Initialize $\mathcal{A}_{\delta}$ as an empty FSA, $visited \gets \{\}$
    \STATE Add node $t$ to $\mathcal{A}_{\delta}$
    \STATE \textsc{ExtractOne}($t$, $\mathcal{A}_{\delta}$, $visited$)
    \STATE \COMMENT{ Merge the partial result into the final result }
    \STATE $\mathcal{A}_R^{\text{ctx}} \gets$ \textsc{FSAUnion}($\mathcal{A}_R^{\text{ctx}}$, $\mathcal{A}_{\delta}$)
\ENDFOR

\vspace{1em}

\FUNCTION{\textsc{ExtractOne}($start$, $\mathcal{A}_{\delta}$, $visited$)}
    \IF{$start$ in $visited$}
        \STATE \textbf{return}
    \ENDIF
    \STATE Add $start$ to $visited$
    \STATE \COMMENT{ Stop search for nodes with rule-referencing edges }
    \IF{$start$ is a final node in $\mathcal{P}$ \textbf{or} \\ \hspace{4.8ex} has an edge referencing another rule}
        \STATE Mark $start$ as final in $\mathcal{A}_{\delta}$
        \STATE \textbf{return}
    \ENDIF
    \STATE \COMMENT{Now all outward edges of $start$ are character edges}
    \FOR{edge $start \xrightarrow{c} end$ from $start$}
        \STATE Add $end$ and $start \xrightarrow{c} end$ to $\mathcal{A}_{\delta}$
        \STATE \textsc{ExtractOne}($end$, $\mathcal{A}_{\delta}$, $visited$)
    \ENDFOR
\ENDFUNCTION

\end{algorithmic}
\end{algorithm}

Although the adaptive token mask cache effectively reduces the number of tokens checked at runtime, checking all context-dependent tokens remains an efficiency bottleneck at runtime. To further reduce the number of context-dependent tokens, \xg{} introduces context expansion, which leverages the grammar's context information to reject more context-dependent tokens during preprocessing, as shown in \figref{fig:ctx_exp}.

As described in the last section, a token is context-dependent when we reach the end of the current rule during matching, but there is still a remaining part of the token that requires further checking by returning to the parent rules. However, through analysis of the grammar, we can observe that in a large portion of the cases, the remaining part of a token is invalid. This is because, for many rules, when they reach their end, there are only a limited number of positions they can return to within their parent rules, and the set of strings that can be further matched from those positions is also limited.

Based on this observation, context expansion precomputes, for each rule, the set of strings that can be accepted after returning to the parent rules, called the \emph{expanded suffix}. The remaining unmatched part of the context-dependent tokens should either be a prefix of
the expanded suffix or start with the expanded suffix. Otherwise, they are rejected. This filtering process effectively reduces the number of context-dependent tokens by eliminating those that would fail in higher-level rule contexts. Applied to the Llama-3.1 model and JSON grammar, this technique reduces context-dependent tokens by 90\% (from 1,134 to 120).

\algoref{alg:context_expansion} describes the context expansion process that finds the expanded suffix of each rule. For a rule $R$, we utilize a finite state automaton~(FSA) $\mathcal{A}_R^{\text{ctx}}$ (ctx is the abbreviation for context) to represent the expanded suffix, and that is extracted from the pushdown automata. We first find all edges $e = (s, t)$ in the pushdown automata that references $R$ and belongs to rule $R'$. $R'$ is not necessarily different from $R$. Then we find a subgraph of the automaton of rule $R'$ starting from $t$ to represent the possible strings that can follow $R$ via depth-first search (DFS). However, we will not consider edges in the subgraph that reference other rules to avoid recursive references between rules, so the edges in the extracted subgraph will only have character labels. If a node has both character edges and edges referencing other rules, we will stop the search at this node. The extracted subgraph is then merged into $\mathcal{A}_R^{\text{ctx}}$. This process is repeated for all rules, and the extracted $\mathcal{A}_R^{\text{ctx}}$ is used to reject context-dependent tokens cannot match any string in it after finishing matching rule $R$.

Although we do not consider rule-referencing edges when extracting the expanded context automata, this algorithm can still extract many useful context information. That is because the inlining optimization introduced in \secref{sec:automata_optim} inlines fragment rules into their parent rules, reducing the need to check into child rules to reject context-dependent tokens.

\subsection{Persistent Execution Stack}
\label{sec:persistent_stack}

\begin{figure}[t]
    \centering
    \includegraphics[width=1\columnwidth]{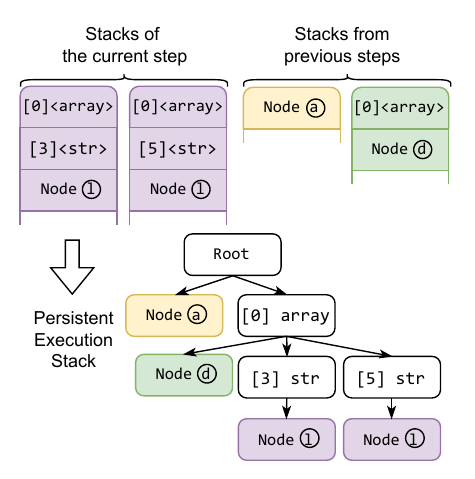}
    \caption{The persistent stack organizes multiple matching stacks from the current step, as well as stacks from previous steps, into a single tree. It reduces memory consumption and supports rolling the state back to previous steps.}
    \label{fig:persist_stack}
\end{figure}

As the grammar engine still needs to handle context-dependent tokens, we need to efficiently execute the pushdown automata for these tokens. Additionally, we also need to execute the pushdown automata for preprocessing the context-independent token sets for all positions in the pushdown automata. In both cases, we need to maintain multiple parallel stacks and branch out as we match the characters in each token. To support efficient state branching, we introduce the persistent execution stack \cite{DRISCOLL198986} to manage the multiple stacks and efficiently execute the pushdown automata. It can also manage the stacks from previous time points and enable the state rollback operation, effectively speeding up the execution of the pushdown automata on a set of tokens.

As shown in \figref{fig:persist_stack}, the persistent execution stack manages a set of stacks, which are either the parallel stacks from the current time point or the stacks from previous time points, into a single tree, and every stack is represented by a path from the root node on the tree. The stack top node is stored as a pointer to the node in the tree. Since the stacks from adjacent time points often share most of the deeper elements and only a few nodes are pushed or popped, this merging avoids memory redundancy for storing multiple stacks. When matching a new character from a token, we may need to split the stack into multiple stacks due to the ambiguity of the grammar, each corresponding to a different expansion of grammar rules. In this case, we only need to split the branch for that stack instead of copying the whole stack, which reduces the overhead of state branching.

Additionally, the persistent execution stack enables fast state rollback by maintaining the stack from previous time points. At runtime, a sliding window of history is maintained. To roll back to a previous state, we only need to change the current stack pointers, which requires constant time. This rollback operation is particularly useful for checking a large set of tokens, as many tokens share a common prefix with other tokens, such as \lstinline{read}, \lstinline{ready}, and \lstinline{reader} all sharing the prefix \lstinline{read}. All the checked tokens are sorted in lexicographical order to find the maximum length of the common prefixes. Then the tokens are checked one by one, and before checking each token, the state rolls back to just after the common prefix with the previous token. Therefore, we can avoid the redundant checks of these common prefixes, reducing the number of characters that need to be checked. For Llama-3.1 model and JSON grammar, this approach reduces the number of characters that need to be checked across the entire vocabulary to 30\%, significantly speeding up the preprocessing stage.

\myparagraph{The rollback operation enables more applications with efficient structured generation.} There are many LLM applications that involve rolling back the output to a previous token. For instance, the jump-forward decoding \cite{yin2024fastjson} requires retokenization, which involves rolling back some tokens in the context and then inserting new tokens. To ensure structured generation can continue after rolling back tokens, we can roll back the automaton state simultaneously with the output token rollback. There are also many LLM applications that requires LLMs generate in a tree structure, such as in Tree-of-thought \cite{10.5555/3666122.3666639}, SGLang \cite{zheng2024sglangefficientexecutionstructured}, and the speculative model in the speculative decoding algorithm SpecInfer \cite{Miao_2024}. We can maintain the automata state for every branch of the output tree, and when the output branches, we can quickly split the automaton state, maintaining separate matching states for each output branch. This branching is fast because we only need to maintain the stack top pointer on the tree for every branch. Therefore, the persistent execution stack enables us to ensure efficient structured generation for all these applications.

\subsection{Pushdown Automata Structure Optimizations}

\label{sec:automata_optim}

We will perform additional optimizations to improve the structure of pushdown automata to speed up the efficiency of final execution. These optimizations draw from traditional compiler optimization concepts, but we find them particularly useful for efficient constrained decoding.

\myparagraph{Rule inlining.}

There could be many fragment rules, i.e. rules with only a few elements, in the specified context-free grammar, which are then converted into small FSA in the pushdown automaton. On the one hand, this increases the ambiguity of the grammar since we need to inspect into these fragment rules and check during the execution of the pushdown automata. On the other hands, during context expansion, references to fragment rules are not considered, so the extracted context automata will be smaller. We will miss the opportunity to reject context-dependent tokens based on the structure of these fragment rules.

To address this issue, we introduce an automatic inlining strategy \cite{10.1145/359810.359830} for fragment rules. We iteratively pick rules that do not reference other rules and inline them into the parent rules. To avoid the explosion of the automaton size, we limit the size of the inlined rule and the size of inlined result to constants. This inlining process almost eliminated fragment rules, thereby improving the efficiency of token checking and enhancing the effectiveness of the context expansion.

\myparagraph{Pushdown automata node merging.}

For pushdown automata, in many cases, the ambiguity comes from multiple outward edges of a node with the same label. When matching tokens, if we arrive at this node, and the next character just matches the label, the matching stack will be split into multiple stacks, one for each outward edge. The increase in the number of stacks increases the computation as we need to check the context-dependent tokens for each stack and merge the token masks. To reduce this kind of ambiguity, the node merging algorithm merges the subsequent nodes that satisfy: a) they are pointed to by edges with the same label originating from the same point b) they are not pointed to by other edges.

Additionally, the epsilon edge also increases the ambiguity of the matching process. An epsilon edge $s \xrightarrow{\epsilon} t$ in the automata means that the matching process can directly move from $s$ to $t$ without consuming any characters. If the matching process arrives at $s$, the execution stack will split into two stacks, one with $s$ at the top and the other with $t$, both of which can continue matching. To reduce this kind of ambiguity, the node merging algorithm also merges the nodes $s$ and $t$ into a single node, as long as $s$ has no other outward edge or $t$ has no zero inward edge.

These two optimizations preserves the equivalence of the automaton, but reduces the number of nodes and edges. At runtime, the number of stacks and the computation required for token checking are reduced, speeding up the mask generation process.

\subsection{Overlapping Mask Generation and LLM Inference}
\label{sec:overlapping}

\begin{figure}[t]
    \centering
    \includegraphics[width=\columnwidth]{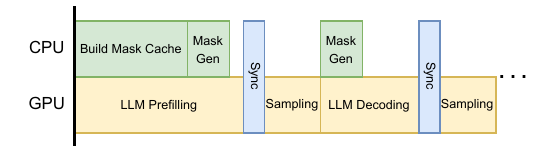}
    \caption{Overlapping building the mask cache with LLM prefilling, and mask generation with LLM decoding to minimize the overhead.}
    \label{fig:pipeline}
\end{figure}

With the optimizations mentioned above, the token mask generation process is significantly accelerated, but it still requires CPU computation. To further eliminate the overhead of constrained decoding, we overlap the computation for mask generation with the LLM inference process, as shown in \figref{fig:pipeline}. We observed that the mask generation process and LLM inference process can be overlapped. That is because the mask generation only requires CPU, and only depends on the previously generated tokens. The LLM inference process except the sampling stage only requires GPU, and also only depends on the previously generated tokens. Therefore, we can parallelize the mask generation process on the CPU with the LLM inference process on the GPU. We will synchronize before sampling, and the GPU will obtain the mask from the CPU and perform masked sampling to generate the new token. Additionally, the preprocessing stage can also be overlapped with the LLM prefilling stage, where the LLM processes the prompt. This orchestration between CPU and GPU ensures that the token restrictions are applied seamlessly, with almost zero overhead for LLM inference. In practice, the time for mask generation is less than the time for LLM inference, so the mask generation process will not become the bottleneck of the generation process.

\section{Evaluation}

We implement \xg in 12,000 lines of core C++ code, and we provide Python bindings to facilitate seamless integration with LLM inference frameworks. In this section, we evaluate \xg to answer the following questions:

\begin{itemize}
\item Can \xg efficiently support each step of constrained decoding? (\secref{sec:unit_test})
\item Does \xg achieve minimal overhead for end-to-end structured generation in LLM serving? (\secref{sec:e2e})
\item How effective is each optimization technique introduced in \xg? (\secref{sec:ablation})
\item How does \xg effect downstream structured generation tasks? (\secref{sec:downstream})
\end{itemize}

\subsection{Mask generation efficiency}
\label{sec:unit_test}

This section evaluates the efficiency of mask generation to measure the overhead introduced by constraint decoding. We first assess regex-based methods using JSON schemas, which can be converted into regex. To test more complex cases beyond regex capabilities, we evaluate context-free grammars, including unconstrained JSON (from ECMA-404~\cite{ECMA-404}), XML (based on the XML 1.0 standard~\cite{xml10}), and a Python DSL (adapted from the Python Grammar Specification~\cite{python_grammar}). Unconstrained JSON cannot be handled by regex-based methods due to its support for arbitrarily nested lists and objects. The Python DSL covers basic control flow (if, for, while) and data types (str, int, float, bool) but ignores indentation. For JSON schema and unconstrained JSON, we use the JSON-mode-eval dataset~\cite{huggingfaceNousResearchjsonmodeevalDatasets}, and for XML and Python, we use a synthetic dataset. For baselines, we choose three popular constrained generation libraries: Outlines~\cite{willard2023efficient} (v1.0), the grammar engine in llama.cpp~\cite{githubGitHubGgerganovllamacpp} (b3998), and lm-format-enforcer~\cite{gat2025lmformat} (v0.10.9, a regex-based method that does not support CFG). All methods are evaluated on Llama-3.1-8B-Instruct using an AMD Ryzen 9 7950X CPU and an NVIDIA RTX 4090 GPU.

\begin{figure}[t]
    \centering
    \includegraphics[width=0.9\columnwidth]{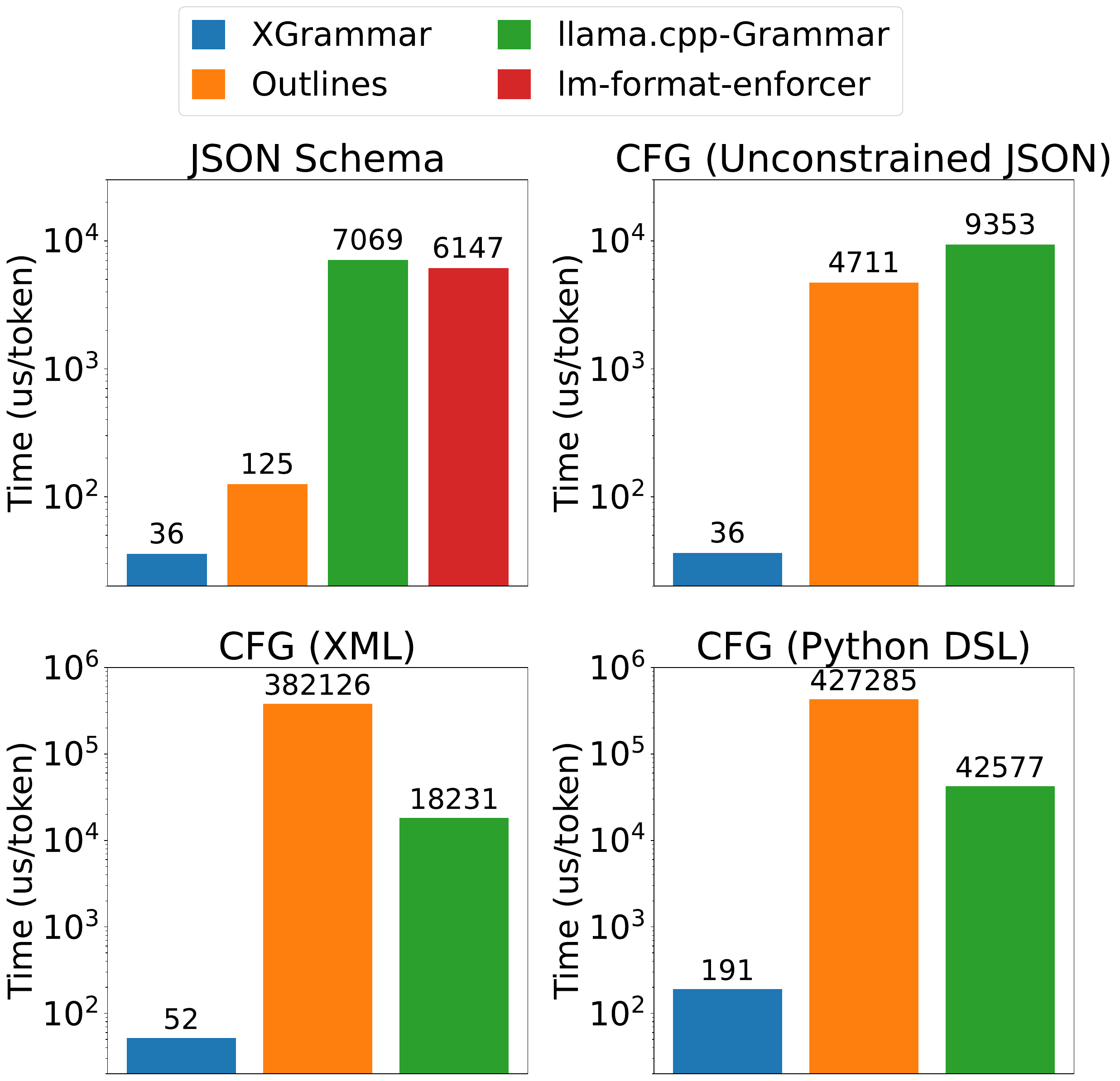}
    \caption{Per token mask generation latency. \xg consistently outperforms existing constrained decoding libraries.}
    \label{fig:eval-unittest}
\end{figure}

The results are shown in Figure~\ref{fig:eval-unittest}. \xg consistently achieves the lowest latency across all tasks, with under 40 µs per token for JSON Schema and CFG (JSON), and under 200 µs for XML and Python DSL. It delivers up to 3x speedup on JSON Schema and over 100x on CFG, compared to the best baseline in each case.

\subsection{End-to-End LLM Engine Evaluation}
\label{sec:e2e}

\begin{figure*}[!t]
    \centering
    \includegraphics[width=0.84\linewidth]{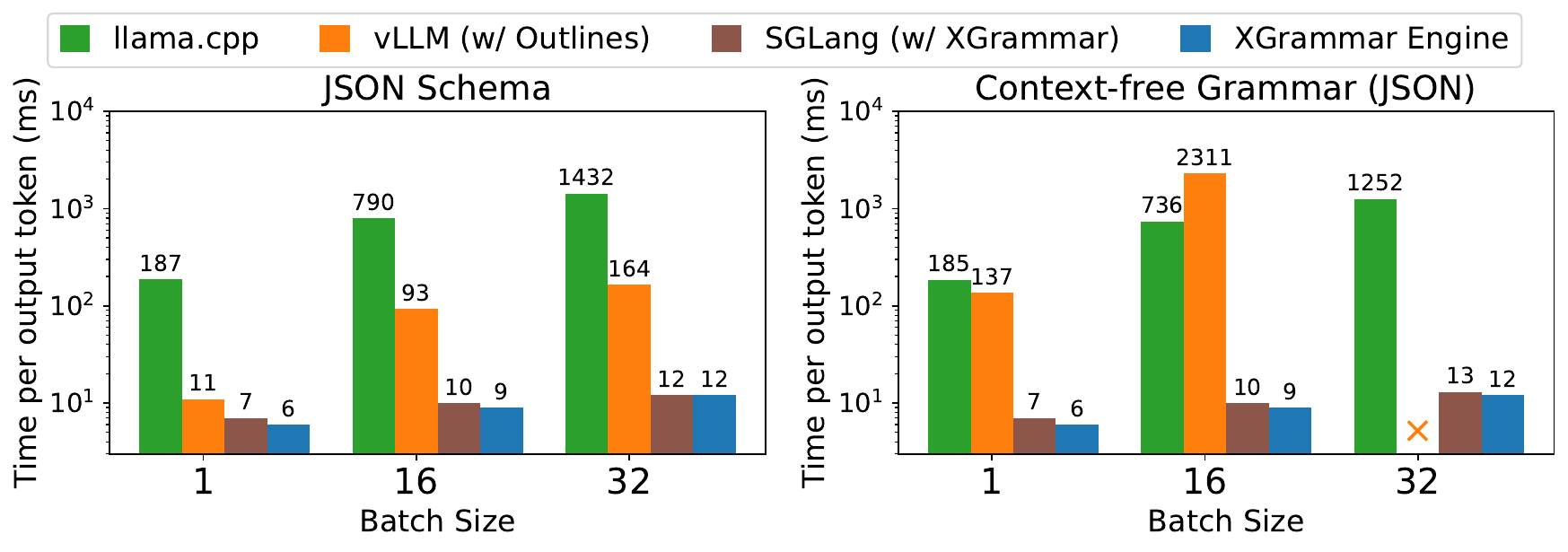}
    \caption{End-to-end evaluation on Llama 3.1 inference with structured constraints. Some results with a batch size of 32 are not reported because their API call time exceeded the API timeout limit of 600 seconds.}
    \label{fig:eval-e2e}
\end{figure*}

\begin{table}[t]
\centering
\caption{End-to-end structured generation efficiency across different models, measured in time per output token (ms) on the JSON Schema task and the Llama-3.1 8B model. \xg demonstrates superior performance across different models.}
\vspace{0.2cm}
\label{tbl:sglang-model-comparison}
\begin{tabular}{|c|c|c|}
\hline
\multirow{2}{*}{\textbf{Model}} & \textbf{SGLang} & \textbf{SGLang} \\
& \textbf{+ Outlines} & \textbf{+ \xg{}} \\
\hline
Llama-3.1 8B & 44.2 & 6.8 \\ \hline
DeepSeek-V2-Lite & \multirow{2}{*}{15.8} & \multirow{2}{*}{4.8} \\
16B MOE & & \\ \hline
\end{tabular}
\end{table}

This section evaluates \xg in LLM serving scenarios. We co-design \xg with the serving engines using the overlapping technique introduced in \secref{sec:overlapping}. We integrate it into widely used end-to-end LLM serving engines, including the C++-based MLC-LLM~\cite{mlc-llm} and the Python-based SGLang~\cite{zheng2024sglangefficientexecutionstructured}, showcasing \xg's adaptability and efficiency across different deployment environments.

We first compare the efficiency of several LLM engines that support structured generation, including vLLM~\cite{kwon2023efficientmemorymanagementlarge} (v0.6.3) with Outlines and llama.cpp with its built-in grammar engine. Efficiency is measured by the average time per output token (TPOT), which reflects the overhead of applying constraints during token generation. The evaluations are conducted using Llama-3.1-8B-Instruct under both JSON schema and CFG (unconstrained JSON). All tests are run in an online serving setting with fixed batch sizes, on hardware with an AMD EPYC 7R13 CPU and an NVIDIA H100 GPU. To ensure a fair comparison, we set a fixed maximum output length. On average, the input has 139 tokens, and the generated output has 53 tokens.

The experiment results are shown in \figref{fig:eval-e2e}. \xg achieves the best TPOT among all baselines for both JSON Schema and CFG. The computation of vLLM and llama.cpp is hindered by their grammar engines' longer preprocessing and per-token processing time. The decrease in TPOT speed in vLLM becomes particularly noticeable with larger batch sizes. This is because a larger batch size leads to higher throughput, putting greater pressure on grammar processing on the CPU side. Overall the \xg-based structured generation solutions can bring up to 80x output token rate compared to existing solutions. This proves the effectiveness of \xg's system optimizations and its co-design with the serving engine.

We also study the effectiveness of XGrammar across different models in end-to-end serving scenarios. As the result shown in \tableref{tbl:sglang-model-comparison}, for various models, SGLang integrated with XGrammar consistently outperforms its integration with Outlines, demonstrating the robustness of \xg's effectiveness across model architectures. This experiment also provides direct evidence that \xg outperforms other constrained decoding library when running on the same serving engine.

Additionally, we examine the overhead of \xg's constrained decoding in end-to-end scenarios by measuring performance on the MLC-LLM engine with and without XGrammar. As shown in \tableref{tbl:grammar-overhead}, enabling XGrammar enhances output quality with nearly zero overhead in TPOT. This is attributed to efficient mask generation and the overlapping of grammar processing with GPU execution.

\begin{table}[t]
    \centering
    \caption{The impact on performance of enabling and disabling XGrammar tested on the MLC-LLM Engine and the Llama-3.1 8B model. TPOT (ms) is reported. XGrammar introduces minimal overhead to the serving engine with better generation quality.}
    \label{tbl:grammar-overhead}
    \vspace{0.2cm}
    \begin{tabular}{|c|c|c|c|}
    \hline
    \multirow{2}{*}{\textbf{Task}} & \textbf{Batch} & \textbf{TPOT w/o} & \textbf{TPOT w/} \\
    & \textbf{Size} & \textbf{XGrammar} & \textbf{XGrammar} \\
    \hline
    \multirow{2}{*}{JSON Schema} & 1 & 6.2 & 6.3 \\ \cline{2-4}
        & 16 & 9.0 & 9.2 \\ \hline
    \multirow{2}{*}{CFG (JSON)} & 1 & 6.3 & 6.3 \\ \cline{2-4}
        & 16 & 9.0 & 9.1 \\ \hline
    \end{tabular}
\end{table}

\subsection{Ablation Study of Optimization Techniques}
\label{sec:ablation}

\begin{table}[t]
    \centering
    \caption{Ablation study of optimization techniques in XGrammar.}
    \label{tbl:ablation}
    \vspace{0.2cm}
    \begin{tabular}{l l}
    \hline
    \textbf{Optimization} & \textbf{Per-token latency (ms)} \\
    \hline
    PDA Baseline & 65.776 \\
    + Node merging & 38.280 (1.7×↓) \\
    + Adaptive token mask cache & 0.154 (248.6×↓) \\
    + Rule inlining & 0.035 (4.4×↓) \\
    + Context Expansion & 0.018 (1.9×↓) \\
    \hline
    \end{tabular}
\end{table}

In this section, we investigate the impact of various optimizations introduced in xgrammar on mask generation performance, to better illustrate our design decisions. We begin by implementing a baseline using a pushdown automaton parser without any optimizations, where each token mask is generated by checking the entire vocabulary to determine whether parsing can proceed. Building on this baseline, we progressively add the optimizations described in this paper, namely node merging, the adaptive token mask cache, rule inlining, and context expansion. For each configuration, we measure the average mask generation time on the CFG (unconstrained JSON) task with the Llama-3.1 8B model and the json-mode-eval dataset. As shown in \tableref{tbl:ablation}, the results show that the adaptive token mask cache has the greatest impact in terms of speedup, while other techniques, including node merging, rule inlining, and context expansion, also yield noticeable improvements.

\subsection{Impact of XGrammar on Structured Generation Tasks}
\label{sec:downstream}

\xg can improve the generation quality of LLMs by ensuring that the output strictly adheres to the given format. We evaluate the impact of XGrammar on two strcutured generation tasks: function calling (i.e., JSON generation guided by a JSON schema) and XML code generation. For function calling, we use the json-mode-eval dataset, while for XML code generation, we rely on a synthetic dataset. We measure the syntactic correctness of the generated function calling outputs and XML code using LLaMA-3.1 8B. As shown in \tableref{tbl:grammar-impact}, \xg significantly improves generation accuracy. We observe that without XGrammar, the model often includes additional explanations alongside the intended code output, or the generated JSON contains an unexpected type. This makes the output unsuitable for direct use by downstream applications. XGrammar avoids this issue by enforcing grammar constraints.

\begin{table}[t]
    \centering
    \caption{Impact of XGrammar on structured generation tasks. XGrammar ensures 100\% syntactic correctness of the generated outputs.}
    \label{tbl:grammar-impact}
    \vspace{0.2cm}
    \begin{tabular}{|l|c|c|}
    \hline
    \multirow{2}{*}{\textbf{Task}} & \textbf{Accuracy w/o} & \textbf{Accuracy w/} \\
    & \textbf{XGrammar} & \textbf{XGrammar} \\
    \hline
    Function calling & 62\% & 100\% \\
    \hline
    XML code generation & 80\% & 100\% \\
    \hline
    \end{tabular}
\end{table}

\section{Related Work}

Several works looked at algorithm improvements for structured generation. \cite{koo2024automatabasedconstraintslanguagemodel} proposes an algorithm to convert character-level pushdown automata to token-level pushdown automata. \cite{wang2023grammarpromptingdomainspecificlanguage} specifies LLM output structure through prompting. \cite{ rozière2024codellamaopenfoundation, codealpaca, li2023starcodersourceyou} explore finetuning LLMs for higher quality structured generation. \xg's approach is orthogonal to these methods and can be combined with these approaches.

There has also been some previous work focusing on constrained decoding. Some methods use regex to represent syntax, such as lm-format-enforcer~\cite{githubGitHubNoamgatlmformatenforcer}, but they cannot handle more complex CFGs. Synchromesh~\cite{poesia2022synchromesh} and llama.cpp~\cite{githubGitHubGgerganovllamacpp} utilizes LR parser and PDA respectively to handle CFG and generate the mask, but requires checking the entire vocabulary at runtime, which incurs large overhead. Outlines~\cite{willard2023efficient} handles the grammar through a lexer and parser, and uses caching to accelerate mask generation, but it only considers the most recent lexer token, which may lead to incorrect judgments when LLM tokens span across multiple lexer tokens (Appendix A in~\cite{koo2024automatabasedconstraintslanguagemodel}). Syncode~\cite{ugare2024syncodellmgenerationgrammar} uses a lexer and parser with a cache spanning multiple lexer tokens, but requires all tokens to be processed offline, leading to substantial preprocessing overhead. In this work, \xg leverages a PDA to support CFGs and introduces an adaptive token mask cache to adaptively handle tokens during preprocessing and runtime, combining other system optimizations to achieve minimal overhead in both stages.

Guidance~\cite{githubGitHubGuidanceaiguidance}, LMQL \cite{githubGitHubEthsrilmql}, SGLang~\cite{zheng2024sglangefficientexecutionstructured}  provide flexible ways to declare the structures. \xg{} is complementary to these improvements and can be used as the backend engine to speedup their execution.

LLM serving engines~\cite{mlc-llm,zheng2024sglangefficientexecutionstructured,kwon2023efficient,hiworldwzj2024ModelTCLightLLM} employ various techniques to support efficient LLM generation for multiple concurrent users, including engine-level techniques such as continuous batching~\cite{orca} for dyanmic request scheduling, and low-level KV cache technique PagedKVCache~\cite{kwon2023efficient} for efficient memory management. Also, AICI~\cite{Moskal2024} proposes a CPU-GPU parallel computation paradigm to accelerate structured generation. These LLM serving engines can leverage \xg{} for efficient, structured generation on top of their existing LLM inference techniques.

\section{Conclusion}

We proposed \xg, a flexible and efficient structured generation engine for LLMs. \xg separates the vocabulary into context-independent tokens and context-dependent ones. It prechecks the context-dependent tokens and stores the result in an adaptive token mask cache. We further introduce a persistent stack to speed up the execution of context-dependent checks. Finally, we co-design the grammar engine with LLM inference to overlap grammar execution with GPU computation. Our system greatly speeds up the token mask generation process in token mask and enables zero overhead structure generation in end-to-end LLM inference flows. We hope our system can enable a broader range of structure generation across platforms.

\section*{Acknowledgements}

This work was supported in part by NSF award CNS-2211882, and gifts from OctoAI, Qualcomm, and CMU opensource software fellowships. We are grateful to Sasa Misailovic for his guidance as the shepherd of this paper. We also appreciate the valuable feedback and discussions from the DeepSeek, SGLang, TensorRT-LLM, vLLM, and WebLLM teams (listed alphabetically). Additionally, we thank Weihua Du, Haoran Peng, Xinyu Yang, Zihao Ye, Jieyu Zhang, Zhihao Zhang, and Ligeng Zhu for their insightful input and thoughtful conversations.



\bibliography{references}

\begin{thebibliography}{52}
\providecommand{\natexlab}[1]{#1}
\providecommand{\url}[1]{\texttt{#1}}
\expandafter\ifx\csname urlstyle\endcsname\relax
  \providecommand{\doi}[1]{doi: #1}\else
  \providecommand{\doi}{doi: \begingroup \urlstyle{rm}\Url}\fi

\bibitem[Beurer-Kellner(2023)]{githubGitHubEthsrilmql}
Beurer-Kellner, L.
\newblock {G}it{H}ub - eth-sri/lmql: {A} language for constraint-guided and efficient {L}{L}{M} programming. --- github.com.
\newblock \url{https://github.com/eth-sri/lmql}, 2023.
\newblock [Accessed 31-10-2024].

\bibitem[Bray et~al.(2008)Bray, Paoli, Sperberg-McQueen, Maler, and Yergeau]{xml10}
Bray, T., Paoli, J., Sperberg-McQueen, C.~M., Maler, E., and Yergeau, F.
\newblock {Extensible Markup Language (XML) 1.0 (Fifth Edition)}.
\newblock \url{https://www.w3.org/TR/xml}, November 2008.
\newblock W3C Recommendation, 26 November 2008.

\bibitem[Bridle(1989)]{NIPS1989_0336dcba}
Bridle, J.
\newblock Training stochastic model recognition algorithms as networks can lead to maximum mutual information estimation of parameters.
\newblock In Touretzky, D. (ed.), \emph{Advances in Neural Information Processing Systems}, volume~2. Morgan-Kaufmann, 1989.
\newblock URL \url{https://proceedings.neurips.cc/paper_files/paper/1989/file/0336dcbab05b9d5ad24f4333c7658a0e-Paper.pdf}.

\bibitem[Chaudhary(2023)]{codealpaca}
Chaudhary, S.
\newblock Code alpaca: An instruction-following llama model for code generation.
\newblock \url{https://github.com/sahil280114/codealpaca}, 2023.

\bibitem[Chen et~al.(2021)Chen, Tworek, Jun, Yuan, de~Oliveira~Pinto, Kaplan, Edwards, Burda, Joseph, Brockman, Ray, Puri, Krueger, Petrov, Khlaaf, Sastry, Mishkin, Chan, Gray, Ryder, Pavlov, Power, Kaiser, Bavarian, Winter, Tillet, Such, Cummings, Plappert, Chantzis, Barnes, Herbert-Voss, Guss, Nichol, Paino, Tezak, Tang, Babuschkin, Balaji, Jain, Saunders, Hesse, Carr, Leike, Achiam, Misra, Morikawa, Radford, Knight, Brundage, Murati, Mayer, Welinder, McGrew, Amodei, McCandlish, Sutskever, and Zaremba]{chen2021evaluatinglargelanguagemodels}
Chen, M., Tworek, J., Jun, H., Yuan, Q., de~Oliveira~Pinto, H.~P., Kaplan, J., Edwards, H., Burda, Y., Joseph, N., Brockman, G., Ray, A., Puri, R., Krueger, G., Petrov, M., Khlaaf, H., Sastry, G., Mishkin, P., Chan, B., Gray, S., Ryder, N., Pavlov, M., Power, A., Kaiser, L., Bavarian, M., Winter, C., Tillet, P., Such, F.~P., Cummings, D., Plappert, M., Chantzis, F., Barnes, E., Herbert-Voss, A., Guss, W.~H., Nichol, A., Paino, A., Tezak, N., Tang, J., Babuschkin, I., Balaji, S., Jain, S., Saunders, W., Hesse, C., Carr, A.~N., Leike, J., Achiam, J., Misra, V., Morikawa, E., Radford, A., Knight, M., Brundage, M., Murati, M., Mayer, K., Welinder, P., McGrew, B., Amodei, D., McCandlish, S., Sutskever, I., and Zaremba, W.
\newblock Evaluating large language models trained on code, 2021.
\newblock URL \url{https://arxiv.org/abs/2107.03374}.

\bibitem[Chomsky(1956)]{1056813}
Chomsky, N.
\newblock Three models for the description of language.
\newblock \emph{IRE Transactions on Information Theory}, 2\penalty0 (3):\penalty0 113--124, 1956.
\newblock \doi{10.1109/TIT.1956.1056813}.

\bibitem[Deutsch et~al.(2019)Deutsch, Upadhyay, and Roth]{deutsch-etal-2019-general}
Deutsch, D., Upadhyay, S., and Roth, D.
\newblock A general-purpose algorithm for constrained sequential inference.
\newblock In Bansal, M. and Villavicencio, A. (eds.), \emph{Proceedings of the 23rd Conference on Computational Natural Language Learning (CoNLL)}, pp.\  482--492, Hong Kong, China, November 2019. Association for Computational Linguistics.
\newblock \doi{10.18653/v1/K19-1045}.
\newblock URL \url{https://aclanthology.org/K19-1045}.

\bibitem[Driscoll et~al.(1989)Driscoll, Sarnak, Sleator, and Tarjan]{DRISCOLL198986}
Driscoll, J.~R., Sarnak, N., Sleator, D.~D., and Tarjan, R.~E.
\newblock Making data structures persistent.
\newblock \emph{Journal of Computer and System Sciences}, 38\penalty0 (1):\penalty0 86--124, 1989.
\newblock ISSN 0022-0000.
\newblock \doi{https://doi.org/10.1016/0022-0000(89)90034-2}.
\newblock URL \url{https://www.sciencedirect.com/science/article/pii/0022000089900342}.

\bibitem[Dubey et~al.(2024{\natexlab{a}})Dubey, Jauhri, Pandey, Kadian, Al-Dahle, Letman, Mathur, Schelten, Yang, Fan, Goyal, Hartshorn, Yang, Mitra, Sravankumar, Korenev, Hinsvark, Rao, Zhang, Rodriguez, Gregerson, Spataru, Roziere, Biron, Tang, Chern, Caucheteux, Nayak, Bi, Marra, McConnell, Keller, Touret, Wu, Wong, Ferrer, Nikolaidis, Allonsius, Song, Pintz, Livshits, Esiobu, Choudhary, Mahajan, Garcia-Olano, Perino, Hupkes, Lakomkin, AlBadawy, Lobanova, Dinan, Smith, Radenovic, Zhang, Synnaeve, Lee, Anderson, Nail, Mialon, Pang, Cucurell, Nguyen, Korevaar, Xu, Touvron, Zarov, Ibarra, Kloumann, Misra, Evtimov, Copet, Lee, Geffert, Vranes, Park, Mahadeokar, Shah, van~der Linde, Billock, Hong, Lee, Fu, Chi, Huang, Liu, Wang, Yu, Bitton, Spisak, Park, Rocca, Johnstun, Saxe, Jia, Alwala, Upasani, Plawiak, Li, Heafield, Stone, El-Arini, Iyer, Malik, Chiu, Bhalla, Rantala-Yeary, van~der Maaten, Chen, Tan, Jenkins, Martin, Madaan, Malo, Blecher, Landzaat, de~Oliveira, Muzzi, Pasupuleti, Singh, Paluri, Kardas,
  Oldham, Rita, Pavlova, Kambadur, Lewis, Si, Singh, Hassan, Goyal, Torabi, Bashlykov, Bogoychev, Chatterji, Duchenne, Çelebi, Alrassy, Zhang, Li, Vasic, Weng, Bhargava, Dubal, Krishnan, Koura, Xu, He, Dong, Srinivasan, Ganapathy, Calderer, Cabral, Stojnic, Raileanu, Girdhar, Patel, Sauvestre, Polidoro, Sumbaly, Taylor, Silva, Hou, Wang, Hosseini, Chennabasappa, Singh, Bell, Kim, Edunov, Nie, Narang, Raparthy, Shen, Wan, Bhosale, Zhang, Vandenhende, Batra, Whitman, Sootla, Collot, Gururangan, Borodinsky, Herman, Fowler, Sheasha, Georgiou, Scialom, Speckbacher, Mihaylov, Xiao, Karn, Goswami, Gupta, Ramanathan, Kerkez, Gonguet, Do, Vogeti, Petrovic, Chu, Xiong, Fu, Meers, Martinet, Wang, Tan, Xie, Jia, Wang, Goldschlag, Gaur, Babaei, Wen, Song, Zhang, Li, Mao, Coudert, Yan, Chen, Papakipos, Singh, Grattafiori, Jain, Kelsey, Shajnfeld, Gangidi, Victoria, Goldstand, Menon, Sharma, Boesenberg, Vaughan, Baevski, Feinstein, Kallet, Sangani, Yunus, Lupu, Alvarado, Caples, Gu, Ho, Poulton, Ryan, Ramchandani, Franco,
  Saraf, Chowdhury, Gabriel, Bharambe, Eisenman, Yazdan, James, Maurer, Leonhardi, Huang, Loyd, Paola, Paranjape, Liu, Wu, Ni, Hancock, Wasti, Spence, Stojkovic, Gamido, Montalvo, Parker, Burton, Mejia, Wang, Kim, Zhou, Hu, Chu, Cai, Tindal, Feichtenhofer, Civin, Beaty, Kreymer, Li, Wyatt, Adkins, Xu, Testuggine, David, Parikh, Liskovich, Foss, Wang, Le, Holland, Dowling, Jamil, Montgomery, Presani, Hahn, Wood, Brinkman, Arcaute, Dunbar, Smothers, Sun, Kreuk, Tian, Ozgenel, Caggioni, Guzmán, Kanayet, Seide, Florez, Schwarz, Badeer, Swee, Halpern, Thattai, Herman, Sizov, Guangyi, Zhang, Lakshminarayanan, Shojanazeri, Zou, Wang, Zha, Habeeb, Rudolph, Suk, Aspegren, Goldman, Damlaj, Molybog, Tufanov, Veliche, Gat, Weissman, Geboski, Kohli, Asher, Gaya, Marcus, Tang, Chan, Zhen, Reizenstein, Teboul, Zhong, Jin, Yang, Cummings, Carvill, Shepard, McPhie, Torres, Ginsburg, Wang, Wu, U, Saxena, Prasad, Khandelwal, Zand, Matosich, Veeraraghavan, Michelena, Li, Huang, Chawla, Lakhotia, Huang, Chen, Garg, A, Silva,
  Bell, Zhang, Guo, Yu, Moshkovich, Wehrstedt, Khabsa, Avalani, Bhatt, Tsimpoukelli, Mankus, Hasson, Lennie, Reso, Groshev, Naumov, Lathi, Keneally, Seltzer, Valko, Restrepo, Patel, Vyatskov, Samvelyan, Clark, Macey, Wang, Hermoso, Metanat, Rastegari, Bansal, Santhanam, Parks, White, Bawa, Singhal, Egebo, Usunier, Laptev, Dong, Zhang, Cheng, Chernoguz, Hart, Salpekar, Kalinli, Kent, Parekh, Saab, Balaji, Rittner, Bontrager, Roux, Dollar, Zvyagina, Ratanchandani, Yuvraj, Liang, Alao, Rodriguez, Ayub, Murthy, Nayani, Mitra, Li, Hogan, Battey, Wang, Maheswari, Howes, Rinott, Bondu, Datta, Chugh, Hunt, Dhillon, Sidorov, Pan, Verma, Yamamoto, Ramaswamy, Lindsay, Lindsay, Feng, Lin, Zha, Shankar, Zhang, Zhang, Wang, Agarwal, Sajuyigbe, Chintala, Max, Chen, Kehoe, Satterfield, Govindaprasad, Gupta, Cho, Virk, Subramanian, Choudhury, Goldman, Remez, Glaser, Best, Kohler, Robinson, Li, Zhang, Matthews, Chou, Shaked, Vontimitta, Ajayi, Montanez, Mohan, Kumar, Mangla, Albiero, Ionescu, Poenaru, Mihailescu, Ivanov, Li,
  Wang, Jiang, Bouaziz, Constable, Tang, Wang, Wu, Wang, Xia, Wu, Gao, Chen, Hu, Jia, Qi, Li, Zhang, Zhang, Adi, Nam, Yu, Wang, Hao, Qian, He, Rait, DeVito, Rosnbrick, Wen, Yang, and Zhao]{dubey2024llama3herdmodels}
Dubey, A., Jauhri, A., Pandey, A., Kadian, A., Al-Dahle, A., Letman, A., Mathur, A., Schelten, A., Yang, A., Fan, A., Goyal, A., Hartshorn, A., Yang, A., Mitra, A., Sravankumar, A., Korenev, A., Hinsvark, A., Rao, A., Zhang, A., Rodriguez, A., Gregerson, A., Spataru, A., Roziere, B., Biron, B., Tang, B., Chern, B., Caucheteux, C., Nayak, C., Bi, C., Marra, C., McConnell, C., Keller, C., Touret, C., Wu, C., Wong, C., Ferrer, C.~C., Nikolaidis, C., Allonsius, D., Song, D., Pintz, D., Livshits, D., Esiobu, D., Choudhary, D., Mahajan, D., Garcia-Olano, D., Perino, D., Hupkes, D., Lakomkin, E., AlBadawy, E., Lobanova, E., Dinan, E., Smith, E.~M., Radenovic, F., Zhang, F., Synnaeve, G., Lee, G., Anderson, G.~L., Nail, G., Mialon, G., Pang, G., Cucurell, G., Nguyen, H., Korevaar, H., Xu, H., Touvron, H., Zarov, I., Ibarra, I.~A., Kloumann, I., Misra, I., Evtimov, I., Copet, J., Lee, J., Geffert, J., Vranes, J., Park, J., Mahadeokar, J., Shah, J., van~der Linde, J., Billock, J., Hong, J., Lee, J., Fu, J., Chi, J.,
  Huang, J., Liu, J., Wang, J., Yu, J., Bitton, J., Spisak, J., Park, J., Rocca, J., Johnstun, J., Saxe, J., Jia, J., Alwala, K.~V., Upasani, K., Plawiak, K., Li, K., Heafield, K., Stone, K., El-Arini, K., Iyer, K., Malik, K., Chiu, K., Bhalla, K., Rantala-Yeary, L., van~der Maaten, L., Chen, L., Tan, L., Jenkins, L., Martin, L., Madaan, L., Malo, L., Blecher, L., Landzaat, L., de~Oliveira, L., Muzzi, M., Pasupuleti, M., Singh, M., Paluri, M., Kardas, M., Oldham, M., Rita, M., Pavlova, M., Kambadur, M., Lewis, M., Si, M., Singh, M.~K., Hassan, M., Goyal, N., Torabi, N., Bashlykov, N., Bogoychev, N., Chatterji, N., Duchenne, O., Çelebi, O., Alrassy, P., Zhang, P., Li, P., Vasic, P., Weng, P., Bhargava, P., Dubal, P., Krishnan, P., Koura, P.~S., Xu, P., He, Q., Dong, Q., Srinivasan, R., Ganapathy, R., Calderer, R., Cabral, R.~S., Stojnic, R., Raileanu, R., Girdhar, R., Patel, R., Sauvestre, R., Polidoro, R., Sumbaly, R., Taylor, R., Silva, R., Hou, R., Wang, R., Hosseini, S., Chennabasappa, S., Singh, S.,
  Bell, S., Kim, S.~S., Edunov, S., Nie, S., Narang, S., Raparthy, S., Shen, S., Wan, S., Bhosale, S., Zhang, S., Vandenhende, S., Batra, S., Whitman, S., Sootla, S., Collot, S., Gururangan, S., Borodinsky, S., Herman, T., Fowler, T., Sheasha, T., Georgiou, T., Scialom, T., Speckbacher, T., Mihaylov, T., Xiao, T., Karn, U., Goswami, V., Gupta, V., Ramanathan, V., Kerkez, V., Gonguet, V., Do, V., Vogeti, V., Petrovic, V., Chu, W., Xiong, W., Fu, W., Meers, W., Martinet, X., Wang, X., Tan, X.~E., Xie, X., Jia, X., Wang, X., Goldschlag, Y., Gaur, Y., Babaei, Y., Wen, Y., Song, Y., Zhang, Y., Li, Y., Mao, Y., Coudert, Z.~D., Yan, Z., Chen, Z., Papakipos, Z., Singh, A., Grattafiori, A., Jain, A., Kelsey, A., Shajnfeld, A., Gangidi, A., Victoria, A., Goldstand, A., Menon, A., Sharma, A., Boesenberg, A., Vaughan, A., Baevski, A., Feinstein, A., Kallet, A., Sangani, A., Yunus, A., Lupu, A., Alvarado, A., Caples, A., Gu, A., Ho, A., Poulton, A., Ryan, A., Ramchandani, A., Franco, A., Saraf, A., Chowdhury, A., Gabriel,
  A., Bharambe, A., Eisenman, A., Yazdan, A., James, B., Maurer, B., Leonhardi, B., Huang, B., Loyd, B., Paola, B.~D., Paranjape, B., Liu, B., Wu, B., Ni, B., Hancock, B., Wasti, B., Spence, B., Stojkovic, B., Gamido, B., Montalvo, B., Parker, C., Burton, C., Mejia, C., Wang, C., Kim, C., Zhou, C., Hu, C., Chu, C.-H., Cai, C., Tindal, C., Feichtenhofer, C., Civin, D., Beaty, D., Kreymer, D., Li, D., Wyatt, D., Adkins, D., Xu, D., Testuggine, D., David, D., Parikh, D., Liskovich, D., Foss, D., Wang, D., Le, D., Holland, D., Dowling, E., Jamil, E., Montgomery, E., Presani, E., Hahn, E., Wood, E., Brinkman, E., Arcaute, E., Dunbar, E., Smothers, E., Sun, F., Kreuk, F., Tian, F., Ozgenel, F., Caggioni, F., Guzmán, F., Kanayet, F., Seide, F., Florez, G.~M., Schwarz, G., Badeer, G., Swee, G., Halpern, G., Thattai, G., Herman, G., Sizov, G., Guangyi, Zhang, Lakshminarayanan, G., Shojanazeri, H., Zou, H., Wang, H., Zha, H., Habeeb, H., Rudolph, H., Suk, H., Aspegren, H., Goldman, H., Damlaj, I., Molybog, I.,
  Tufanov, I., Veliche, I.-E., Gat, I., Weissman, J., Geboski, J., Kohli, J., Asher, J., Gaya, J.-B., Marcus, J., Tang, J., Chan, J., Zhen, J., Reizenstein, J., Teboul, J., Zhong, J., Jin, J., Yang, J., Cummings, J., Carvill, J., Shepard, J., McPhie, J., Torres, J., Ginsburg, J., Wang, J., Wu, K., U, K.~H., Saxena, K., Prasad, K., Khandelwal, K., Zand, K., Matosich, K., Veeraraghavan, K., Michelena, K., Li, K., Huang, K., Chawla, K., Lakhotia, K., Huang, K., Chen, L., Garg, L., A, L., Silva, L., Bell, L., Zhang, L., Guo, L., Yu, L., Moshkovich, L., Wehrstedt, L., Khabsa, M., Avalani, M., Bhatt, M., Tsimpoukelli, M., Mankus, M., Hasson, M., Lennie, M., Reso, M., Groshev, M., Naumov, M., Lathi, M., Keneally, M., Seltzer, M.~L., Valko, M., Restrepo, M., Patel, M., Vyatskov, M., Samvelyan, M., Clark, M., Macey, M., Wang, M., Hermoso, M.~J., Metanat, M., Rastegari, M., Bansal, M., Santhanam, N., Parks, N., White, N., Bawa, N., Singhal, N., Egebo, N., Usunier, N., Laptev, N.~P., Dong, N., Zhang, N., Cheng, N.,
  Chernoguz, O., Hart, O., Salpekar, O., Kalinli, O., Kent, P., Parekh, P., Saab, P., Balaji, P., Rittner, P., Bontrager, P., Roux, P., Dollar, P., Zvyagina, P., Ratanchandani, P., Yuvraj, P., Liang, Q., Alao, R., Rodriguez, R., Ayub, R., Murthy, R., Nayani, R., Mitra, R., Li, R., Hogan, R., Battey, R., Wang, R., Maheswari, R., Howes, R., Rinott, R., Bondu, S.~J., Datta, S., Chugh, S., Hunt, S., Dhillon, S., Sidorov, S., Pan, S., Verma, S., Yamamoto, S., Ramaswamy, S., Lindsay, S., Lindsay, S., Feng, S., Lin, S., Zha, S.~C., Shankar, S., Zhang, S., Zhang, S., Wang, S., Agarwal, S., Sajuyigbe, S., Chintala, S., Max, S., Chen, S., Kehoe, S., Satterfield, S., Govindaprasad, S., Gupta, S., Cho, S., Virk, S., Subramanian, S., Choudhury, S., Goldman, S., Remez, T., Glaser, T., Best, T., Kohler, T., Robinson, T., Li, T., Zhang, T., Matthews, T., Chou, T., Shaked, T., Vontimitta, V., Ajayi, V., Montanez, V., Mohan, V., Kumar, V.~S., Mangla, V., Albiero, V., Ionescu, V., Poenaru, V., Mihailescu, V.~T., Ivanov, V., Li,
  W., Wang, W., Jiang, W., Bouaziz, W., Constable, W., Tang, X., Wang, X., Wu, X., Wang, X., Xia, X., Wu, X., Gao, X., Chen, Y., Hu, Y., Jia, Y., Qi, Y., Li, Y., Zhang, Y., Zhang, Y., Adi, Y., Nam, Y., Yu, Wang, Hao, Y., Qian, Y., He, Y., Rait, Z., DeVito, Z., Rosnbrick, Z., Wen, Z., Yang, Z., and Zhao, Z.
\newblock The llama 3 herd of models, 2024{\natexlab{a}}.
\newblock URL \url{https://arxiv.org/abs/2407.21783}.

\bibitem[Dubey et~al.(2024{\natexlab{b}})Dubey, Jauhri, Pandey, Kadian, Al-Dahle, Letman, Mathur, Schelten, Yang, Fan, et~al.]{dubey2024llama}
Dubey, A., Jauhri, A., Pandey, A., Kadian, A., Al-Dahle, A., Letman, A., Mathur, A., Schelten, A., Yang, A., Fan, A., et~al.
\newblock The llama 3 herd of models.
\newblock \emph{arXiv preprint arXiv:2407.21783}, 2024{\natexlab{b}}.

\bibitem[{Ecma International}(2013)]{ECMA-404}
{Ecma International}.
\newblock {ECMA-404 The JSON Data Interchange Standard}.
\newblock Online, 2013.
\newblock \url{https://www.ecma-international.org/publications-and-standards/standards/ecma-404/}.

\bibitem[Evey(1963)]{evey1963theory}
Evey, R.
\newblock \emph{The Theory and Applications of Pushdown Store Machines}.
\newblock Mathematical linguistic and automatic translation: Report to National Science Foundation. Harvard University, 1963.
\newblock URL \url{https://books.google.com/books?id=mg4yAAAAIAAJ}.

\bibitem[Gat(2024)]{githubGitHubNoamgatlmformatenforcer}
Gat, N.
\newblock {G}it{H}ub - noamgat/lm-format-enforcer: {E}nforce the output format ({J}{S}{O}{N} {S}chema, {R}egex etc) of a language model --- github.com.
\newblock \url{https://github.com/noamgat/lm-format-enforcer}, 2024.
\newblock [Accessed 31-10-2024].

\bibitem[Gat et~al.(2025)]{gat2025lmformat}
Gat, N. et~al.
\newblock lm-format-enforcer.
\newblock \url{https://github.com/noamgat/lm-format-enforcer}, 2025.
\newblock Accessed: 2025-03-27.

\bibitem[Gerganov(2023)]{githubGitHubGgerganovllamacpp}
Gerganov, G.
\newblock {G}it{H}ub - ggerganov/llama.cpp: {L}{L}{M} inference in {C}/{C}++ --- github.com.
\newblock \url{https://github.com/ggerganov/llama.cpp}, 2023.
\newblock [Accessed 31-10-2024].

\bibitem[Guidance-ai(2024)]{githubGitHubGuidanceaiguidance}
Guidance-ai.
\newblock {G}it{H}ub - guidance-ai/guidance: {A} guidance language for controlling large language models. --- github.com.
\newblock \url{https://github.com/guidance-ai/guidance}, 2024.
\newblock [Accessed 31-10-2024].

\bibitem[Haas et~al.(2017)Haas, Rossberg, Schuff, Titzer, Holman, Gohman, Wagner, Zakai, and Bastien]{haas2017webassembly}
Haas, A., Rossberg, A., Schuff, D.~L., Titzer, B.~L., Holman, M., Gohman, D., Wagner, L., Zakai, A., and Bastien, J.
\newblock Bringing the web up to speed with webassembly.
\newblock In \emph{Proceedings of the 38th ACM SIGPLAN Conference on Programming Language Design and Implementation}, pp.\  185--200, 2017.

\bibitem[hiworldwzj et~al.(2024)hiworldwzj, shihaobai, sufubao, WANDY666, FlyingFlame, llehtahw, LiangLiu, wxd000000, fuheaven, XHPlus, Chielo, Yong, and\_gate, sangchengmeng, wangzhihong, singularity, Yang, SiYu, Tracin, Granger, Husain, R, SunXiaoye, Peng, Uranus, Bai, Fan, bingo, liuhuakai, and XFPlus]{hiworldwzj2024ModelTCLightLLM}
hiworldwzj, shihaobai, sufubao, WANDY666, FlyingFlame, llehtahw, LiangLiu, wxd000000, fuheaven, XHPlus, Chielo, Yong, Y., and\_gate, sangchengmeng, wangzhihong, singularity, Yang, S., SiYu, W., Tracin, Granger, E., Husain, H., R, S. A. G.~A., SunXiaoye, Peng, T., Uranus, Bai, Y., Fan, Y., bingo, liuhuakai, and XFPlus.
\newblock \emph{ModelTC/lightllm}.
\newblock 10 2024.
\newblock URL \url{https://github.com/ModelTC/lightllm}.

\bibitem[Jiang et~al.(2023)Jiang, Sablayrolles, Mensch, Bamford, Chaplot, de~las Casas, Bressand, Lengyel, Lample, Saulnier, Lavaud, Lachaux, Stock, Scao, Lavril, Wang, Lacroix, and Sayed]{jiang2023mistral7b}
Jiang, A.~Q., Sablayrolles, A., Mensch, A., Bamford, C., Chaplot, D.~S., de~las Casas, D., Bressand, F., Lengyel, G., Lample, G., Saulnier, L., Lavaud, L.~R., Lachaux, M.-A., Stock, P., Scao, T.~L., Lavril, T., Wang, T., Lacroix, T., and Sayed, W.~E.
\newblock Mistral 7b, 2023.
\newblock URL \url{https://arxiv.org/abs/2310.06825}.

\bibitem[Koo et~al.(2024)Koo, Liu, and He]{koo2024automatabasedconstraintslanguagemodel}
Koo, T., Liu, F., and He, L.
\newblock Automata-based constraints for language model decoding, 2024.
\newblock URL \url{https://arxiv.org/abs/2407.08103}.

\bibitem[Kuchnik et~al.(2023)Kuchnik, Smith, and Amvrosiadis]{kuchnik2023validating}
Kuchnik, M., Smith, V., and Amvrosiadis, G.
\newblock Validating large language models with relm.
\newblock \emph{Proceedings of Machine Learning and Systems}, 5:\penalty0 457--476, 2023.

\bibitem[Kwon et~al.(2023{\natexlab{a}})Kwon, Li, Zhuang, Sheng, Zheng, Yu, Gonzalez, Zhang, and Stoica]{kwon2023efficient}
Kwon, W., Li, Z., Zhuang, S., Sheng, Y., Zheng, L., Yu, C.~H., Gonzalez, J.~E., Zhang, H., and Stoica, I.
\newblock Efficient memory management for large language model serving with pagedattention.
\newblock In \emph{Proceedings of the ACM SIGOPS 29th Symposium on Operating Systems Principles}, 2023{\natexlab{a}}.

\bibitem[Kwon et~al.(2023{\natexlab{b}})Kwon, Li, Zhuang, Sheng, Zheng, Yu, Gonzalez, Zhang, and Stoica]{kwon2023efficientmemorymanagementlarge}
Kwon, W., Li, Z., Zhuang, S., Sheng, Y., Zheng, L., Yu, C.~H., Gonzalez, J.~E., Zhang, H., and Stoica, I.
\newblock Efficient memory management for large language model serving with pagedattention, 2023{\natexlab{b}}.
\newblock URL \url{https://arxiv.org/abs/2309.06180}.

\bibitem[LangChain(2024)]{langchainToolCalling}
LangChain.
\newblock {T}ool {C}alling with {L}ang{C}hain --- blog.langchain.dev.
\newblock \url{https://blog.langchain.dev/tool-calling-with-langchain/}, 2024.
\newblock [Accessed 26-10-2024].

\bibitem[Li et~al.(2023)Li, Allal, Zi, Muennighoff, Kocetkov, Mou, Marone, Akiki, Li, Chim, Liu, Zheltonozhskii, Zhuo, Wang, Dehaene, Davaadorj, Lamy-Poirier, Monteiro, Shliazhko, Gontier, Meade, Zebaze, Yee, Umapathi, Zhu, Lipkin, Oblokulov, Wang, Murthy, Stillerman, Patel, Abulkhanov, Zocca, Dey, Zhang, Fahmy, Bhattacharyya, Yu, Singh, Luccioni, Villegas, Kunakov, Zhdanov, Romero, Lee, Timor, Ding, Schlesinger, Schoelkopf, Ebert, Dao, Mishra, Gu, Robinson, Anderson, Dolan-Gavitt, Contractor, Reddy, Fried, Bahdanau, Jernite, Ferrandis, Hughes, Wolf, Guha, von Werra, and de~Vries]{li2023starcodersourceyou}
Li, R., Allal, L.~B., Zi, Y., Muennighoff, N., Kocetkov, D., Mou, C., Marone, M., Akiki, C., Li, J., Chim, J., Liu, Q., Zheltonozhskii, E., Zhuo, T.~Y., Wang, T., Dehaene, O., Davaadorj, M., Lamy-Poirier, J., Monteiro, J., Shliazhko, O., Gontier, N., Meade, N., Zebaze, A., Yee, M.-H., Umapathi, L.~K., Zhu, J., Lipkin, B., Oblokulov, M., Wang, Z., Murthy, R., Stillerman, J., Patel, S.~S., Abulkhanov, D., Zocca, M., Dey, M., Zhang, Z., Fahmy, N., Bhattacharyya, U., Yu, W., Singh, S., Luccioni, S., Villegas, P., Kunakov, M., Zhdanov, F., Romero, M., Lee, T., Timor, N., Ding, J., Schlesinger, C., Schoelkopf, H., Ebert, J., Dao, T., Mishra, M., Gu, A., Robinson, J., Anderson, C.~J., Dolan-Gavitt, B., Contractor, D., Reddy, S., Fried, D., Bahdanau, D., Jernite, Y., Ferrandis, C.~M., Hughes, S., Wolf, T., Guha, A., von Werra, L., and de~Vries, H.
\newblock Starcoder: may the source be with you!, 2023.
\newblock URL \url{https://arxiv.org/abs/2305.06161}.

\bibitem[Liu et~al.(2023)Liu, Jiang, Zhang, Liu, Zhang, Biswas, and Stone]{liu2023llmpempoweringlargelanguage}
Liu, B., Jiang, Y., Zhang, X., Liu, Q., Zhang, S., Biswas, J., and Stone, P.
\newblock Llm+p: Empowering large language models with optimal planning proficiency, 2023.
\newblock URL \url{https://arxiv.org/abs/2304.11477}.

\bibitem[Miao et~al.(2024)Miao, Oliaro, Zhang, Cheng, Wang, Zhang, Wong, Zhu, Yang, Shi, Shi, Chen, Arfeen, Abhyankar, and Jia]{Miao_2024}
Miao, X., Oliaro, G., Zhang, Z., Cheng, X., Wang, Z., Zhang, Z., Wong, R. Y.~Y., Zhu, A., Yang, L., Shi, X., Shi, C., Chen, Z., Arfeen, D., Abhyankar, R., and Jia, Z.
\newblock Specinfer: Accelerating large language model serving with tree-based speculative inference and verification.
\newblock In \emph{Proceedings of the 29th ACM International Conference on Architectural Support for Programming Languages and Operating Systems, Volume 3}, ASPLOS ’24, pp.\  932–949. ACM, April 2024.
\newblock \doi{10.1145/3620666.3651335}.
\newblock URL \url{http://dx.doi.org/10.1145/3620666.3651335}.

\bibitem[{MLC team}(2023{\natexlab{a}})]{mlc-llm}
{MLC team}.
\newblock {MLC-LLM}, 2023{\natexlab{a}}.
\newblock URL \url{https://github.com/mlc-ai/mlc-llm}.

\bibitem[{MLC team}(2023{\natexlab{b}})]{web-llm}
{MLC team}.
\newblock {WebLLM}, 2023{\natexlab{b}}.
\newblock URL \url{https://github.com/mlc-ai/web-llm}.

\bibitem[Moskal et~al.(2024)Moskal, Musuvathi, and {K\i c\i man}]{Moskal2024}
Moskal, M., Musuvathi, M., and {K\i c\i man}, E.
\newblock {AI Controller Interface}.
\newblock \url{https://github.com/microsoft/aici/}, 2024.

\bibitem[Mozannar et~al.(2024)Mozannar, Bansal, Fourney, and Horvitz]{mozannar2024reading}
Mozannar, H., Bansal, G., Fourney, A., and Horvitz, E.
\newblock Reading between the lines: Modeling user behavior and costs in ai-assisted programming.
\newblock In \emph{Proceedings of the CHI Conference on Human Factors in Computing Systems}, pp.\  1--16, 2024.

\bibitem[NousResearch(2024)]{huggingfaceNousResearchjsonmodeevalDatasets}
NousResearch.
\newblock {N}ous{R}esearch/json-mode-eval · {D}atasets at {H}ugging {F}ace --- huggingface.co.
\newblock \url{https://huggingface.co/datasets/NousResearch/json-mode-eval}, 2024.
\newblock [Accessed 31-10-2024].

\bibitem[OpenAI(2024)]{openai2024funccall}
OpenAI.
\newblock {F}unction {C}alling - {O}pen{A}{I} {A}{P}{I}.
\newblock \url{https://platform.openai.com/docs/guides/function-calling}, 2024.
\newblock [Accessed 26-10-2024].

\bibitem[OpenAI et~al.(2024)OpenAI, Achiam, Adler, Agarwal, Ahmad, Akkaya, Aleman, Almeida, Altenschmidt, Altman, Anadkat, Avila, Babuschkin, Balaji, Balcom, Baltescu, Bao, Bavarian, Belgum, Bello, Berdine, Bernadett-Shapiro, Berner, Bogdonoff, Boiko, Boyd, Brakman, Brockman, Brooks, Brundage, Button, Cai, Campbell, Cann, Carey, Carlson, Carmichael, Chan, Chang, Chantzis, Chen, Chen, Chen, Chen, Chen, Chess, Cho, Chu, Chung, Cummings, Currier, Dai, Decareaux, Degry, Deutsch, Deville, Dhar, Dohan, Dowling, Dunning, Ecoffet, Eleti, Eloundou, Farhi, Fedus, Felix, Fishman, Forte, Fulford, Gao, Georges, Gibson, Goel, Gogineni, Goh, Gontijo-Lopes, Gordon, Grafstein, Gray, Greene, Gross, Gu, Guo, Hallacy, Han, Harris, He, Heaton, Heidecke, Hesse, Hickey, Hickey, Hoeschele, Houghton, Hsu, Hu, Hu, Huizinga, Jain, Jain, Jang, Jiang, Jiang, Jin, Jin, Jomoto, Jonn, Jun, Kaftan, Łukasz Kaiser, Kamali, Kanitscheider, Keskar, Khan, Kilpatrick, Kim, Kim, Kim, Kirchner, Kiros, Knight, Kokotajlo, Łukasz Kondraciuk, Kondrich,
  Konstantinidis, Kosic, Krueger, Kuo, Lampe, Lan, Lee, Leike, Leung, Levy, Li, Lim, Lin, Lin, Litwin, Lopez, Lowe, Lue, Makanju, Malfacini, Manning, Markov, Markovski, Martin, Mayer, Mayne, McGrew, McKinney, McLeavey, McMillan, McNeil, Medina, Mehta, Menick, Metz, Mishchenko, Mishkin, Monaco, Morikawa, Mossing, Mu, Murati, Murk, Mély, Nair, Nakano, Nayak, Neelakantan, Ngo, Noh, Ouyang, O'Keefe, Pachocki, Paino, Palermo, Pantuliano, Parascandolo, Parish, Parparita, Passos, Pavlov, Peng, Perelman, de~Avila Belbute~Peres, Petrov, de~Oliveira~Pinto, Michael, Pokorny, Pokrass, Pong, Powell, Power, Power, Proehl, Puri, Radford, Rae, Ramesh, Raymond, Real, Rimbach, Ross, Rotsted, Roussez, Ryder, Saltarelli, Sanders, Santurkar, Sastry, Schmidt, Schnurr, Schulman, Selsam, Sheppard, Sherbakov, Shieh, Shoker, Shyam, Sidor, Sigler, Simens, Sitkin, Slama, Sohl, Sokolowsky, Song, Staudacher, Such, Summers, Sutskever, Tang, Tezak, Thompson, Tillet, Tootoonchian, Tseng, Tuggle, Turley, Tworek, Uribe, Vallone, Vijayvergiya,
  Voss, Wainwright, Wang, Wang, Wang, Ward, Wei, Weinmann, Welihinda, Welinder, Weng, Weng, Wiethoff, Willner, Winter, Wolrich, Wong, Workman, Wu, Wu, Wu, Xiao, Xu, Yoo, Yu, Yuan, Zaremba, Zellers, Zhang, Zhang, Zhao, Zheng, Zhuang, Zhuk, and Zoph]{openai2024gpt4technicalreport}
OpenAI, Achiam, J., Adler, S., Agarwal, S., Ahmad, L., Akkaya, I., Aleman, F.~L., Almeida, D., Altenschmidt, J., Altman, S., Anadkat, S., Avila, R., Babuschkin, I., Balaji, S., Balcom, V., Baltescu, P., Bao, H., Bavarian, M., Belgum, J., Bello, I., Berdine, J., Bernadett-Shapiro, G., Berner, C., Bogdonoff, L., Boiko, O., Boyd, M., Brakman, A.-L., Brockman, G., Brooks, T., Brundage, M., Button, K., Cai, T., Campbell, R., Cann, A., Carey, B., Carlson, C., Carmichael, R., Chan, B., Chang, C., Chantzis, F., Chen, D., Chen, S., Chen, R., Chen, J., Chen, M., Chess, B., Cho, C., Chu, C., Chung, H.~W., Cummings, D., Currier, J., Dai, Y., Decareaux, C., Degry, T., Deutsch, N., Deville, D., Dhar, A., Dohan, D., Dowling, S., Dunning, S., Ecoffet, A., Eleti, A., Eloundou, T., Farhi, D., Fedus, L., Felix, N., Fishman, S.~P., Forte, J., Fulford, I., Gao, L., Georges, E., Gibson, C., Goel, V., Gogineni, T., Goh, G., Gontijo-Lopes, R., Gordon, J., Grafstein, M., Gray, S., Greene, R., Gross, J., Gu, S.~S., Guo, Y., Hallacy,
  C., Han, J., Harris, J., He, Y., Heaton, M., Heidecke, J., Hesse, C., Hickey, A., Hickey, W., Hoeschele, P., Houghton, B., Hsu, K., Hu, S., Hu, X., Huizinga, J., Jain, S., Jain, S., Jang, J., Jiang, A., Jiang, R., Jin, H., Jin, D., Jomoto, S., Jonn, B., Jun, H., Kaftan, T., Łukasz Kaiser, Kamali, A., Kanitscheider, I., Keskar, N.~S., Khan, T., Kilpatrick, L., Kim, J.~W., Kim, C., Kim, Y., Kirchner, J.~H., Kiros, J., Knight, M., Kokotajlo, D., Łukasz Kondraciuk, Kondrich, A., Konstantinidis, A., Kosic, K., Krueger, G., Kuo, V., Lampe, M., Lan, I., Lee, T., Leike, J., Leung, J., Levy, D., Li, C.~M., Lim, R., Lin, M., Lin, S., Litwin, M., Lopez, T., Lowe, R., Lue, P., Makanju, A., Malfacini, K., Manning, S., Markov, T., Markovski, Y., Martin, B., Mayer, K., Mayne, A., McGrew, B., McKinney, S.~M., McLeavey, C., McMillan, P., McNeil, J., Medina, D., Mehta, A., Menick, J., Metz, L., Mishchenko, A., Mishkin, P., Monaco, V., Morikawa, E., Mossing, D., Mu, T., Murati, M., Murk, O., Mély, D., Nair, A., Nakano, R.,
  Nayak, R., Neelakantan, A., Ngo, R., Noh, H., Ouyang, L., O'Keefe, C., Pachocki, J., Paino, A., Palermo, J., Pantuliano, A., Parascandolo, G., Parish, J., Parparita, E., Passos, A., Pavlov, M., Peng, A., Perelman, A., de~Avila Belbute~Peres, F., Petrov, M., de~Oliveira~Pinto, H.~P., Michael, Pokorny, Pokrass, M., Pong, V.~H., Powell, T., Power, A., Power, B., Proehl, E., Puri, R., Radford, A., Rae, J., Ramesh, A., Raymond, C., Real, F., Rimbach, K., Ross, C., Rotsted, B., Roussez, H., Ryder, N., Saltarelli, M., Sanders, T., Santurkar, S., Sastry, G., Schmidt, H., Schnurr, D., Schulman, J., Selsam, D., Sheppard, K., Sherbakov, T., Shieh, J., Shoker, S., Shyam, P., Sidor, S., Sigler, E., Simens, M., Sitkin, J., Slama, K., Sohl, I., Sokolowsky, B., Song, Y., Staudacher, N., Such, F.~P., Summers, N., Sutskever, I., Tang, J., Tezak, N., Thompson, M.~B., Tillet, P., Tootoonchian, A., Tseng, E., Tuggle, P., Turley, N., Tworek, J., Uribe, J. F.~C., Vallone, A., Vijayvergiya, A., Voss, C., Wainwright, C., Wang,
  J.~J., Wang, A., Wang, B., Ward, J., Wei, J., Weinmann, C., Welihinda, A., Welinder, P., Weng, J., Weng, L., Wiethoff, M., Willner, D., Winter, C., Wolrich, S., Wong, H., Workman, L., Wu, S., Wu, J., Wu, M., Xiao, K., Xu, T., Yoo, S., Yu, K., Yuan, Q., Zaremba, W., Zellers, R., Zhang, C., Zhang, M., Zhao, S., Zheng, T., Zhuang, J., Zhuk, W., and Zoph, B.
\newblock Gpt-4 technical report, 2024.
\newblock URL \url{https://arxiv.org/abs/2303.08774}.

\bibitem[Pearce et~al.(2022)Pearce, Tan, Ahmad, Karri, and Dolan-Gavitt]{pearce2022examiningzeroshotvulnerabilityrepair}
Pearce, H., Tan, B., Ahmad, B., Karri, R., and Dolan-Gavitt, B.
\newblock Examining zero-shot vulnerability repair with large language models, 2022.
\newblock URL \url{https://arxiv.org/abs/2112.02125}.

\bibitem[Poesia et~al.(2022)Poesia, Polozov, Le, Tiwari, Soares, Meek, and Gulwani]{poesia2022synchromesh}
Poesia, G., Polozov, O., Le, V., Tiwari, A., Soares, G., Meek, C., and Gulwani, S.
\newblock Synchromesh: Reliable code generation from pre-trained language models.
\newblock \emph{arXiv preprint arXiv:2201.11227}, 2022.

\bibitem[{Python Software Foundation}(2024)]{python_grammar}
{Python Software Foundation}.
\newblock Python full grammar specification.
\newblock \url{https://docs.python.org/3/reference/grammar.html}, 2024.
\newblock Accessed: 2025-03-05.

\bibitem[Rozière et~al.(2024)Rozière, Gehring, Gloeckle, Sootla, Gat, Tan, Adi, Liu, Sauvestre, Remez, Rapin, Kozhevnikov, Evtimov, Bitton, Bhatt, Ferrer, Grattafiori, Xiong, Défossez, Copet, Azhar, Touvron, Martin, Usunier, Scialom, and Synnaeve]{rozière2024codellamaopenfoundation}
Rozière, B., Gehring, J., Gloeckle, F., Sootla, S., Gat, I., Tan, X.~E., Adi, Y., Liu, J., Sauvestre, R., Remez, T., Rapin, J., Kozhevnikov, A., Evtimov, I., Bitton, J., Bhatt, M., Ferrer, C.~C., Grattafiori, A., Xiong, W., Défossez, A., Copet, J., Azhar, F., Touvron, H., Martin, L., Usunier, N., Scialom, T., and Synnaeve, G.
\newblock Code llama: Open foundation models for code, 2024.
\newblock URL \url{https://arxiv.org/abs/2308.12950}.

\bibitem[Scheifler(1977)]{10.1145/359810.359830}
Scheifler, R.~W.
\newblock An analysis of inline substitution for a structured programming language.
\newblock \emph{Commun. ACM}, 20\penalty0 (9):\penalty0 647–654, September 1977.
\newblock ISSN 0001-0782.
\newblock \doi{10.1145/359810.359830}.
\newblock URL \url{https://doi.org/10.1145/359810.359830}.

\bibitem[Scholak et~al.(2021)Scholak, Schucher, and Bahdanau]{scholak-etal-2021-picard}
Scholak, T., Schucher, N., and Bahdanau, D.
\newblock {PICARD}: Parsing incrementally for constrained auto-regressive decoding from language models.
\newblock In Moens, M.-F., Huang, X., Specia, L., and Yih, S. W.-t. (eds.), \emph{Proceedings of the 2021 Conference on Empirical Methods in Natural Language Processing}, pp.\  9895--9901, Online and Punta Cana, Dominican Republic, November 2021. Association for Computational Linguistics.
\newblock \doi{10.18653/v1/2021.emnlp-main.779}.
\newblock URL \url{https://aclanthology.org/2021.emnlp-main.779}.

\bibitem[Schützenberger(1963)]{SCHUTZENBERGER1963246}
Schützenberger, M.
\newblock On context-free languages and push-down automata.
\newblock \emph{Information and Control}, 6\penalty0 (3):\penalty0 246--264, 1963.
\newblock ISSN 0019-9958.
\newblock \doi{https://doi.org/10.1016/S0019-9958(63)90306-1}.
\newblock URL \url{https://www.sciencedirect.com/science/article/pii/S0019995863903061}.

\bibitem[Ugare et~al.(2024)Ugare, Suresh, Kang, Misailovic, and Singh]{ugare2024syncodellmgenerationgrammar}
Ugare, S., Suresh, T., Kang, H., Misailovic, S., and Singh, G.
\newblock Syncode: Llm generation with grammar augmentation, 2024.
\newblock URL \url{https://arxiv.org/abs/2403.01632}.

\bibitem[Wang et~al.(2023)Wang, Wang, Wang, Cao, Saurous, and Kim]{wang2023grammarpromptingdomainspecificlanguage}
Wang, B., Wang, Z., Wang, X., Cao, Y., Saurous, R.~A., and Kim, Y.
\newblock Grammar prompting for domain-specific language generation with large language models, 2023.
\newblock URL \url{https://arxiv.org/abs/2305.19234}.

\bibitem[Wang et~al.(2019)Wang, Cho, and Gu]{wang2019neuralmachinetranslationbytelevel}
Wang, C., Cho, K., and Gu, J.
\newblock Neural machine translation with byte-level subwords, 2019.
\newblock URL \url{https://arxiv.org/abs/1909.03341}.

\bibitem[Wang et~al.(2021)Wang, Wang, Joty, and Hoi]{wang2021codet5identifierawareunifiedpretrained}
Wang, Y., Wang, W., Joty, S., and Hoi, S. C.~H.
\newblock Codet5: Identifier-aware unified pre-trained encoder-decoder models for code understanding and generation, 2021.
\newblock URL \url{https://arxiv.org/abs/2109.00859}.

\bibitem[Willard \& Louf(2023)Willard and Louf]{willard2023efficient}
Willard, B.~T. and Louf, R.
\newblock Efficient guided generation for llms.
\newblock \emph{arXiv preprint arXiv:2307.09702}, 2023.

\bibitem[Yang et~al.(2024)Yang, Yang, Hui, Zheng, Yu, Zhou, Li, Li, Liu, Huang, et~al.]{yang2024qwen2}
Yang, A., Yang, B., Hui, B., Zheng, B., Yu, B., Zhou, C., Li, C., Li, C., Liu, D., Huang, F., et~al.
\newblock Qwen2 technical report.
\newblock \emph{arXiv preprint arXiv:2407.10671}, 2024.

\bibitem[Yao et~al.(2024)Yao, Yu, Zhao, Shafran, Griffiths, Cao, and Narasimhan]{10.5555/3666122.3666639}
Yao, S., Yu, D., Zhao, J., Shafran, I., Griffiths, T.~L., Cao, Y., and Narasimhan, K.
\newblock Tree of thoughts: deliberate problem solving with large language models.
\newblock In \emph{Proceedings of the 37th International Conference on Neural Information Processing Systems}, NIPS '23, Red Hook, NY, USA, 2024. Curran Associates Inc.

\bibitem[Yin et~al.(2024)Yin, Sheng, and Zheng]{yin2024fastjson}
Yin, L., Sheng, Y., and Zheng, L.
\newblock Fast json decoding for local llms with compressed finite state machine.
\newblock \url{https://lmsys.org/blog/2024-02-05-compressed-fsm/}, February 2024.
\newblock Accessed: 2025-03-26.

\bibitem[Yu et~al.(2022)Yu, Jeong, Kim, Kim, and Chun]{orca}
Yu, G.-I., Jeong, J.~S., Kim, G.-W., Kim, S., and Chun, B.-G.
\newblock Orca: A distributed serving system for {Transformer-Based} generative models.
\newblock In \emph{16th USENIX Symposium on Operating Systems Design and Implementation (OSDI 22)}, pp.\  521--538, Carlsbad, CA, July 2022. USENIX Association.
\newblock ISBN 978-1-939133-28-1.
\newblock URL \url{https://www.usenix.org/conference/osdi22/presentation/yu}.

\bibitem[Zakai(2011)]{zakai2011emscripten}
Zakai, A.
\newblock Emscripten: an llvm-to-javascript compiler.
\newblock In \emph{Proceedings of the ACM international conference companion on Object oriented programming systems languages and applications companion}, pp.\  301--312, 2011.

\bibitem[Zheng et~al.(2024)Zheng, Yin, Xie, Sun, Huang, Yu, Cao, Kozyrakis, Stoica, Gonzalez, Barrett, and Sheng]{zheng2024sglangefficientexecutionstructured}
Zheng, L., Yin, L., Xie, Z., Sun, C., Huang, J., Yu, C.~H., Cao, S., Kozyrakis, C., Stoica, I., Gonzalez, J.~E., Barrett, C., and Sheng, Y.
\newblock Sglang: Efficient execution of structured language model programs, 2024.
\newblock URL \url{https://arxiv.org/abs/2312.07104}.

\end{thebibliography}
\bibliographystyle{mlsys2024}

\appendix
\section{Formal Definition of the PDA Variant}
\label{sec:appendix-pda}

In our paper, we define a variant of the pushdown automaton (PDA) that is equivalent to the original definition, but is designed to facilitate the description of the parsing algorithm and the construction of the token mask cache, since the keys of the token mask cache are precisely the states in this PDA. It is defined as the tuple
\[
P = \bigl( R, \Sigma, \{A_r\}_{r \in R}, q_{\text{main}}, \delta \bigr),
\]
where:
\begin{itemize}
  \item $R$ is a finite set of grammar rules.
  \item $\Sigma$ is a finite input alphabet.
  \item For each rule $r \in R$, the corresponding finite state automaton is given by
  \[
  A_r = \bigl( Q_r, \Sigma \cup R, q^{\text{start}}_r, F_r, \delta_r \bigr),
  \]
  where $Q_r$ is a finite set of states, $q^{\text{start}}_r \in Q_r$ is the start state, $F_r \subseteq Q_r$ is the set of accepting states, and $\delta_r$ is the transition function defined over $Q_r$. The transition labels in $A_r$ are drawn from the alphabet $\Sigma \cup R$, which includes both input characters and rule references.
  \item $q_{\text{main}}$ is the start state corresponding to the main rule.
  \item $\delta$ is the global transition function that governs the operation of the PDA by handling two kinds of transitions:
    \begin{itemize}
      \item \emph{Character transitions}: When in a state $q \in Q_r$, reading an input symbol $a \in \Sigma$ may lead to a transition within the same automaton, i.e., $q \xrightarrow{a} q'$.
      \item \emph{Rule reference transitions}: When in a state $q \in Q_r$, a transition labeled by a rule reference to $s \in R$ allows the PDA to push the current return information onto the stack and jump to the start state $q^{\text{start}}_s$ of $A_s$.
    \end{itemize}
\end{itemize}

The parsing state is represented by a set of pairs $\{(s_i, q_i)\}$, where $s_i$ denotes the content of the stack (encoding return information), $q_i$ is the current state (with $q_i \in Q_r$ for some $r \in R$). There can be multiple such pairs because the pushdown automaton can contain non-deterministic transitions, which means there could be multiple possible parsing stacks and states. In the main body of the paper, for simplicity, we place the current state $q_i$ at the top of the stack. Thus, the parsing state is represented as a set of parsing stacks.

We now describe a formal transformation that converts the above variant PDA definition into a standard PDA. To obtain a standard PDA, we construct a new pushdown automaton

\[
P' = \bigl( Q, \Sigma, \Gamma, \delta', q_0, F \bigr)
\]

as follows. The state set $Q$ is defined as the union of all states from the FSAs:
\[
Q = \bigcup_{r \in R} Q_r,
\]
and we set the initial state to be $q_0 = q_{\text{main}}$. The stack alphabet $\Gamma$ is chosen to record return information; we define
\[
\Gamma = \{\, (r, q) \mid r \in R,\; q \in Q_r \,\},
\]
so that each symbol $(r, q)$ encodes the context of a rule call, namely the originating rule and the return state.

The transition function $\delta'$ of the standard PDA is then defined to simulate the behavior of the variant PDA. For each character transition in some $\delta_r$, if
\[
q \xrightarrow{a} q' \quad \text{with } a \in \Sigma \text{ and } q, q' \in Q_r,
\]
we include in $\delta'$ the transition
\[
\delta'(q, a, \gamma) \ni (q', \gamma) \quad \text{for all } \gamma \in \Gamma.
\]
For each rule reference transition in $\delta_r$, if
\[
q \xrightarrow{s} q' \quad \text{with } s \in R,
\]
we simulate the recursive call by defining an $\epsilon$-transition that pushes the return information and transfers control to the called rule. Formally, we set
\[
\delta'(q, \epsilon, \gamma) \ni \Bigl( q^{\text{start}}_s, \, (r,q') \cdot \gamma \Bigr) \quad \text{for all } \gamma \in \Gamma,
\]
where $q$ belongs to the automaton $A_r$, and $(r,q') \cdot \gamma$ denotes the stack obtained by pushing $(r,q')$ onto $\gamma$. Finally, when an automaton $A_r$ reaches an accepting state $q \in F_r$, the standard PDA simulates the return from a recursive call by popping the top of the stack. That is, if the current stack has the form $(r',q') \cdot \gamma$, we define
\[
\delta'(q, \epsilon, (r',q')) \ni (q', \gamma).
\]
The set of accepting states $F$ is defined as those states in $\bigcup_{r \in R} F_r$ that are reached with an empty stack.

This construction shows that character transitions in the variant PDA directly correspond to state transitions without stack operations in $P'$, while rule reference transitions correspond to stack push and jump operations. Similarly, completing the match of an FSA and returning to the parent rule is achieved by a stack pop and a transition to the stored return state. In this way, the standard PDA $P'$ exactly simulates the recursive behavior of our variant PDA.

\section{Synergy between XGrammar and Jump-forward Decoding}
\label{sec:appendix-jump-forward}

Jump-forward decoding is a technique designed to accelerate structured generation. When, based on the current input, the following output can be deterministically inferred from the grammar, it bypasses LLM decoding and sampling by directly tokenizing and appending the output to the context. This improves the end-to-end efficiency of LLM generation. This technique is orthogonal to the constrained decoding adopted by XGrammar. XGrammar further supports jump-forward decoding and demonstrates that the two can be effectively combined to further improve efficiency.

We evaluate performance using the Llama-3.1-8B-Instruct model served through the SGLang engine, running on a machine with an AMD Ryzen 9 7950X CPU and an NVIDIA RTX 4090 GPU. All experiments are conducted with a batch size of 1. We compare our method, XGrammar, against the existing decoding backend Outlines, testing both with and without jump-forward decoding enabled. We measure time per output token as the evaluation metric.

\begin{figure}[t]
  \centering
  \includegraphics[width=0.35\textwidth]{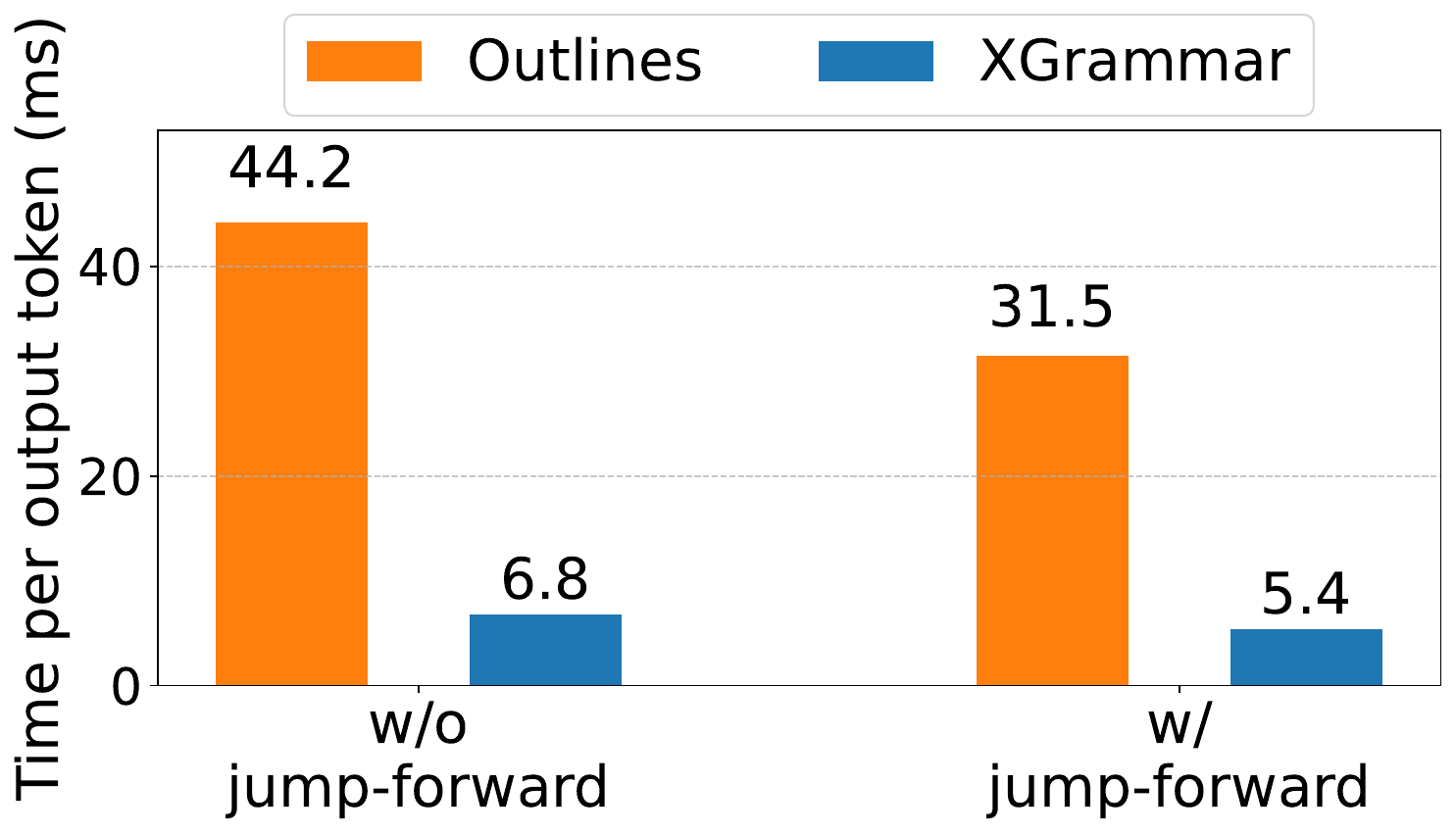}
  \caption{The performance comparison with and without jump-forward decoding on the JSON schema task and the SGLang engine.
  XGrammar combined with jump-forward can achieve optimal TPOT.}
  \label{fig:eval:jump-fwd}
\end{figure}

As shown in \figref{fig:eval:jump-fwd}, XGrammar consistently outperforms Outlines, and achieves the best efficiency when combined with jump-forward decoding—demonstrating its ability to better leverage structural constraints for faster generation.

\section{Cross-platform Deployment of \xg}
\begin{figure}[t]
    \centering
    \includegraphics[width=0.5\textwidth]{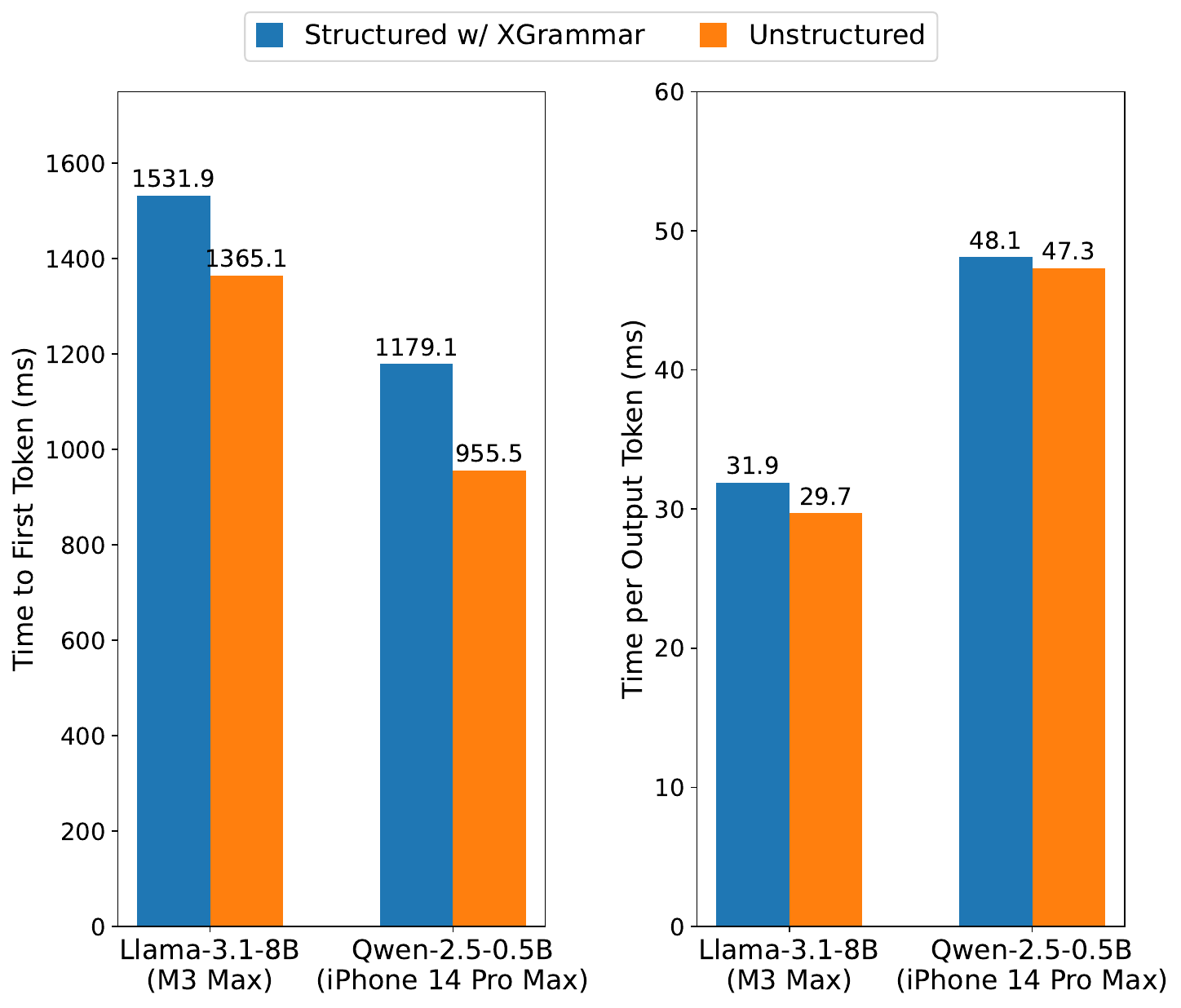}
    \caption{End-to-end performance comparison between structured generation with \xg{} and unstructured generation in browser JavaScript environment.}
    \label{fig:eval:web_eval}
\end{figure}

\label{sec:webllm}

We study bringing \xg{} to a wide variety of platforms.
We leverage Emscripten ~\cite{zakai2011emscripten} to compile
\xg{} into WebAssembly ~\cite{haas2017webassembly} and build a JavaScript binding. This approach enables \xg{} to run in client-side browsers on portable devices like laptops and mobile phones. We further integrate the web-binding with the in-browser LLM inference framework WebLLM ~\cite{web-llm} to enable structured generation.

We evaluate the end-to-end performance with the JSON-mode-eval dataset, using 4-bit quantized models Llama-3.1-8B-Instruct~\cite{dubey2024llama} on a MacBook Pro M3 Max (MacOS 14.5) with Google Chrome, and Qwen-2.5-0.5B-Instruct \cite{yang2024qwen2} on an iPhone 14 Pro Max (iOS 18) with Safari.

The results are shown in \figref{fig:eval:web_eval}. We compare the time to first token (TTFT) and time per output token (TPOT) between structured generation with \xg{} and non-structured generation while ensuring the number of generated tokens is the same. The results show that \xg{} brings close to zero overhead in both settings, enabling a great potential to support future on-device agents with high performance.


\end{document}